\documentclass[9pt]{article}
\usepackage{geometry} 
\usepackage{graphicx}
\graphicspath{ {../figures/} }
\usepackage{caption}
\usepackage{subcaption}
\usepackage[english]{babel}
\usepackage[T1]{fontenc}
\usepackage{ragged2e}
\usepackage{chemformula}
\usepackage{soul}
\usepackage{xcolor}
\usepackage{array}
\usepackage[numbers,sort&compress]{natbib}

\usepackage{hyperref}
\usepackage{changepage} 
\definecolor{lightgreen}{rgb}{0.8,1,0.8}

\definecolor{lightgrey}{rgb}{0.5,0.5,0.5}
\usepackage{colortbl} 
\newcolumntype{C}{>{\centering\arraybackslash}X}  
\definecolor{goodgreen}{rgb}{0.88, 1, 0.88} 

\title{Domain-specific ChatBots for Science using Embeddings}
\author{Kevin G. Yager$^{\ast}$}
\date{}

\begin{document}
{\centering
\null \vspace{3.5 cm}
\noindent \textbf{\huge Domain-specific ChatBots for Science using Embeddings}
\par \null \vspace{1.5 cm}
\noindent \textbf{\Large }
\par \null \vspace{1.5 cm}
\noindent {\Large Kevin G. Yager$^{\ast}$}
\par \null \vspace{0.5 cm}
\noindent {\large Center for Functional Nanomaterials, Brookhaven National Laboratory, Upton, New York 11973, United States}
\par\null \vspace{0.5 cm}
\par\null\par
}

\par

\begin{abstract}
\noindent 
\begin{adjustwidth}{1cm}{1cm} 
Large language models (LLMs) have emerged as powerful machine-learning systems capable of handling a myriad of tasks. Tuned versions of these systems have been turned into chatbots that can respond to user queries on a vast diversity of topics, providing informative and creative replies. However, their application to physical science research remains limited owing to their incomplete knowledge in these areas, contrasted with the needs of rigor and sourcing in science domains. Here, we demonstrate how existing methods and software tools can be easily combined to yield a domain-specific chatbot. The system ingests scientific documents in existing formats, and uses text embedding lookup to provide the LLM with domain-specific contextual information when composing its reply. We similarly demonstrate that existing image embedding methods can be used for search and retrieval across publication figures. These results confirm that LLMs are already suitable for use by physical scientists in accelerating their research efforts.
\end{adjustwidth}
\end{abstract}

\section{Introduction}

Artificial intelligence and machine-learning (AI/ML) methods are growing in sophistication and capability. The application of these methods to the physical sciences is correspondingly seeing enormous growth.\cite{Qiu2016} 
Recent years have seen the convergence of several new trends. Generative AI seeks to create novel outputs that conform to the structure of training data,\cite{9903869, gozalobrizuela2023chatgpt} for instance enabling image synthesis\cite{DALLE2, rombach2021highresolution, Oppenlaender2022} or text generation.
Large language models (LLMs) are generative neural networks trained on text completion, but which can be used for a variety of tasks, including sentiment analysis, code completion, document generation, or for interactive chatbots that respond to users in natural language.\cite{NEURIPS2020_1457c0d6}
The most successful implementations of this concept---such as the generative pre-trained transformer (GPT)\cite{Radford2018}--- exploit the transformer architecture,\cite{vaswani2017attention} which has a self-attention mechanism, allowing the model to weigh the relevance of each input in a sequence and capture the contextual dependencies between words regardless of their distance from each other in the text sequence.
LLMs are part of a general trend in ML towards foundation models---extensive training of large deep neural networks on enormous datasets in a task-agnostic manner.\cite{NEURIPS2020_1457c0d6, foundation2021} 
The performance of LLMs increases with the scale of the training data, network size, and training time. 
There is growing evidence that LLMs are not merely reproducing surface statistics, but are instead learning a meaningful world model.\cite{li2023emergent, akyürek2023learning, kosinski2023theory} Correspondingly, training shows evidence of sudden leaps in performance and corresponding development of surprising new capabilities, suggesting the emergent learning of generalized concepts with more abstraction and sophistication.\cite{Ganguli_2022, wei2022emergent, nanda2023progress, bubeck2023sparks, Webb2023}

Recent work has shown how reinforcement learning using human feedback (RLHF)\cite{ziegler2020finetuning} can be used to further tailor LLMs into generating responses aligned with human desires for helpful and informative text response to user queries. In this way, several efforts have demonstrated high-quality chatbots that can engage in remarkably productive discussion (the most prominent being the ChatGPT system produced by OpenAI). 
These text response systems allow a user to provide input text---which might include instructions, background information, and user question---and solicit a text completion that answers the query. Key engineering aspects of using such systems are managing the finite context window (maximum size available for input text and generated response) and prompt engineering (crafting the input text to elicit the desired behavior). 
The field of LLMs and chat interfaces is advancing rapidly. The prompting process can be elaborated to induce more sophisticated responses akin to deliberation, by using self-analysis of generation quality,\cite{shinn2023reflexion, lightman2023lets} or generating chains of thought through iterative self-prompting.\cite{xu2023reprompting, yao2023tree} 
Chatbots can be augmented with access to external tools through APIs (application programming interfaces),\cite{yao2023react, schick2023toolformer, gao2023pal, liang2023taskmatrixai, shen2023hugginggpt, cai2023large, peng2023check, xu2023rewoo, hsieh2023tool} and can be turned into task-oriented autonomous agents by allowing them to iteratively propose and execute solutions.\cite{shen2023hugginggpt, wang2023voyager, li2023camel, boiko2023emergent}

As these phenomenal capabilities are demonstrated, it is natural to ask how they can be tailored specifically to accelerate research in the physical sciences. 
The most obvious option is to train an LLM on the enormous corpus of scientific publications, thereby producing a chatbot that can converse on science topics. The scale and cost of training an LLM from scratch is daunting, excluding all but the largest research groups. 
Meta AI trained Galactica, an LLM optimized for Science;\cite{taylor2022galactica} however its public availability was short-lived, owing to backlash associated with its propensity for fabricating plausible-sounding but ultimately nonsense scientific text. 
This ``hallucination'' behavior is a key challenge in deploying LLMs, arising from the inherently interpolative nature of neural networks, combined with human preferences selecting for confident answers in the RLHF step. 

Instead of training an LLM from scratch, another option is to fine-tune an existing model on additional domain-specific data. Several efforts have demonstrated highly efficient strategies for performing this step, most notably the low-rank adaptation method,\cite{hu2021lora, dettmers2023qlora} which uses rank decomposition matrices to reduce the number of parameters during retraining. 
Even with such efficiency gains, it remains daunting for the non-expert to deploy, fine-tune, and utilize an LLM locally. In particular, physical scientists typically lack the expertise or inclination to take on such efforts, which suggests that domain-specific chatbots optimized for physical sciences must wait for attention from larger research efforts.

Here, we demonstrate how existing and available tools can be easily chained together to build a domain-adapted chatbot that can discuss scientific topics. 
Our example implementation can take advantage of available scientific documents in the portable document format (PDF), does not require LLM fine-tuning, and addresses the hallucination problem by making document text extractions available to the chatbot through the input prompt. 
A critical aspect of scientific documents is the technical figures contained within. We demonstrate how image embedding methods can be used to find semantically related content among publication figures or image datasets. Together, these demonstrations suggest that domain-specific chatbots can already be easily deployed by any researcher in the physical sciences, and that there is a corresponding opportunity to accelerate the fundamental research enterprise by embracing these new tools.

\section{Results and Discussion}

\subsection{Chatbot design}

\begin{figure*}
\centering
  \includegraphics[width=7.0cm]{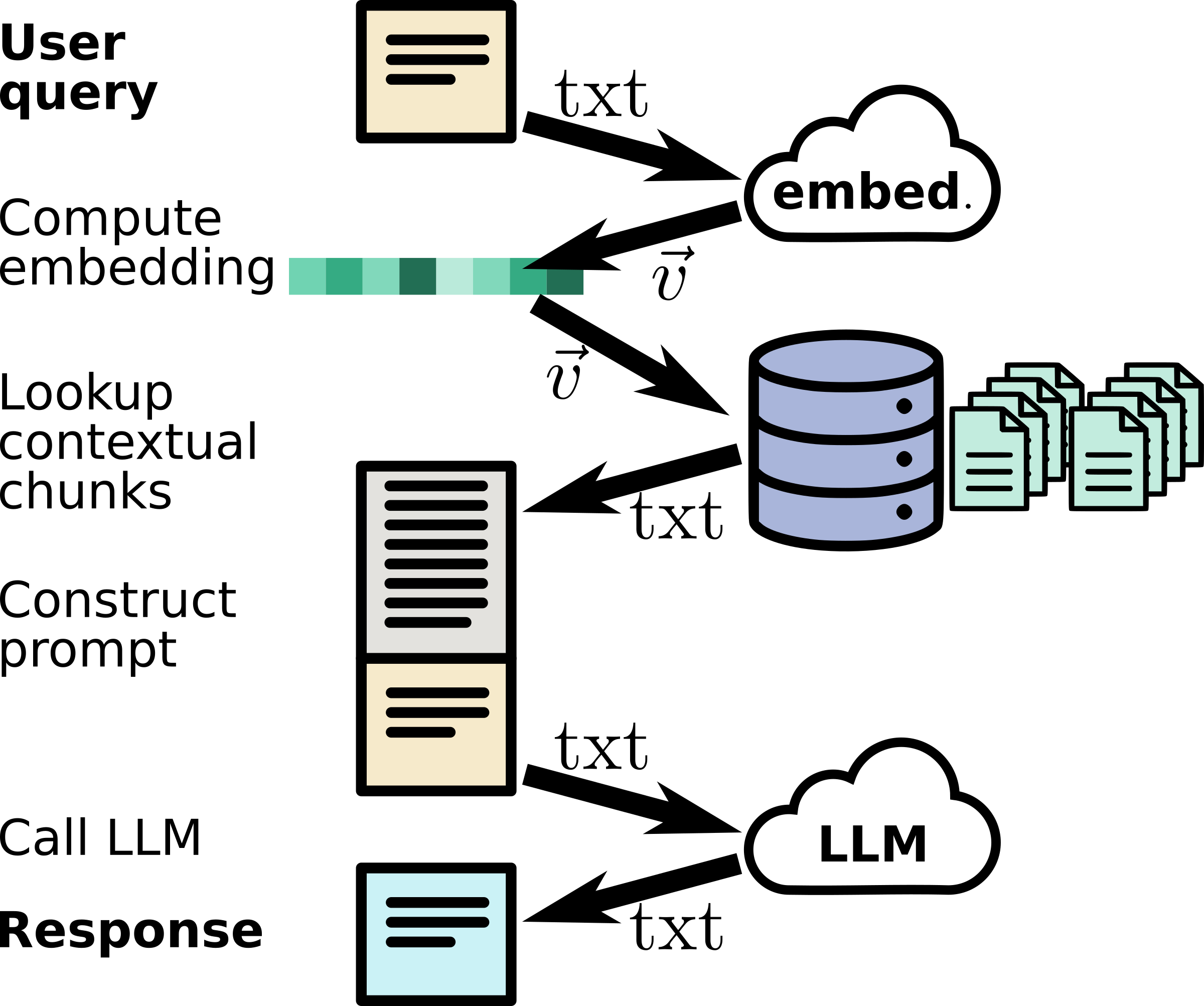}
  \caption{Workflow for generating domain-tailored chatbot response. The user query is first sent to a ML embedding model, which computes an embedding vector ($\vec{v}$) that captures the semantic content of the input. This vector is used to query a pre-computed database of text chunks. Text snippets that are similar to the query (``close'' in the embedding space) are prepended to the user query to construct a prompt. The prompt is sent to a large language model (LLM), which generates a text response for the user.
}
  \label{fgr:prompting}
\end{figure*}

In order to demonstrate the viability of domain-specific chatbots for science topics, we developed a demonstration implementation. Although simple and unrefined, this demo system allows us to investigate utility, and acts as a blueprint for other researchers wishing to deploy similar systems.
The core operating principle of our implementation is to take advantage of text embeddings to retrieve potentially-relevant text extracts (``chunks'') from the corpus of domain-specific documents.\cite{medium2023embeddingex, OpenAI2023embeddingex} 
Text embedding is a natural language processing (NLP) method whereby text is converted into a real-valued vector that encodes the meaning. This conversion is normally performed using a neural network trained to convert text into a concise vector representation, which can then be thought of as a semantic latent space. For instance, words that are close in the embedding space are expected to be similar in meaning. 

A typical LLM chatbot lookup involves constructing an input prompt that involves the user query, where one optionally prepends some additional contextual information (such as the chat history, so that the LLM can assess the context of the most recent user comment). 
In the embedding strategy, we take advantage of the space provided by the context window, adding in text chunks relevant to the query. Procedurally (Figure~\ref{fgr:prompting}), this involves first computing the text embedding of the user query ($q$). 
This embedding vector ($\vec{v_q}$) is compared to the precomputed embeddings across all text chunks (stored in a database). 
Semantically relevant text chunks are identified using the cosine similarity between the user query and each chunk ($\vec{v_c}$):
\begin{align}\label{eq:cosine}
s_{qc} = \frac{\vec{v_q} \cdot \vec{v_c}}{ \|\vec{v_q}\| \|\vec{v_c}\|}
\end{align}
The cosine similarity measures the angle between vectors, and thus assesses whether they point in the same direction in the semantic space. This thus provides a measure of thematic similarity of the two texts being analyzed, as opposed to measuring the similarity in full meaning between the two. 
A small set (5--10) of the most relevant chunks are concatenated and prepended to the user query. This constructed prompt is then sent to the LLM, which generates a coherent response to the query using the available text. The availability of domain-specific text segments allows more specific and meaningful response, including direct quotation and citation of source.

While conceptually simple, this design involves several implementation details that must be considered. 
The first is document format. The version of record for scientific publications tends to be stored in the portable document format (PDF), which is optimized for consistent layout and readability across devices. However, this format entangles content and presentation, making unambiguous extraction of the underlying text quite difficult. While scientific documents of course exist in a more machine-readable format earlier in their development (raw text, latex source, rich text), it is impractical to ask scientists to track down such documents for the myriad of publications relevant to them. Instead, automated conversion of PDF to text is necessary. 
The simplest and most widely used conversion tools (Adobe Acrobat, Grahl PDF Annotator, IntraPDF, PDFTron, 
PDF2Text, etc.) typically do not correctly handle layout, introducing errors such as breaking text within sentences, or mixing between main text and footnotes. The resultant extraction lacks the required coherence. As a result, a series of efforts have arisen to use more sophisticated methods to provide layout-aware document conversion, including ParsCit,\cite{councill-etal-2008-parscit} LA-PDFText,\cite{Ramakrishnan2012} CERMINE,\cite{Tkaczyk2015} OCR++,\cite{singh-etal-2016-ocr} Grobid,\cite{GROBID} and DeepPDF.\cite{DeepPDF} 
We elected to use the Grobid system, which employs ML extraction, provides a clean containerized implementation that acts as server, and converts input PDF files into extensible markup language (XML) outputs that separate the document into meaningful components (title, authors, main text, figures, references, etc.).

\begin{figure*}
\centering
  \includegraphics[width=7.0cm]{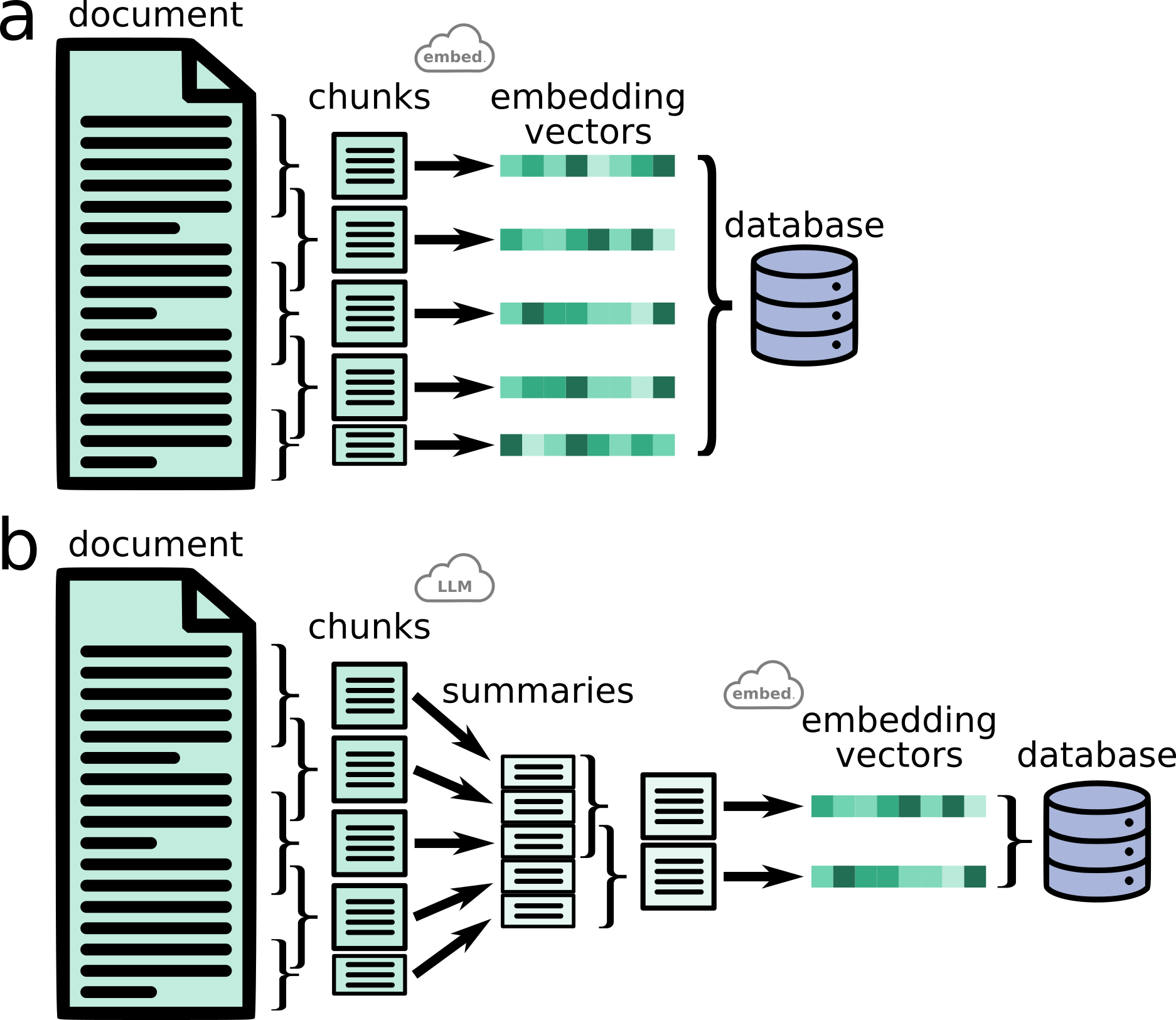}
  \caption{Workflow for ingesting documents for later lookup. (a) Document text is broken up into a set of overlapping chunks. Each is converted into a vector using a text embedding model. The text chunks and corresponding embedding vectors are stored in a database for later retrieval. (b) The raw text chunks can be compressed using an LLM operating as a summarizer. This shorter summary document can be chunked and stored, along with embedding vector, as previously described. These compressed chunks afford the opportunity to avoid redundant information and maximize the information content of the constructed prompt.
}
  \label{fgr:chunking}
\end{figure*}

The structured XML versions of the input documents can be easily parsed, chunked, and stored in a database. The publication title and author list is extracted to compose a concise document name, while the main text is broken into a set of overlapping chunks (Figure~\ref{fgr:chunking}a). While segmenting the text could be performed in a text-aware manner (e.g. by paragraph), breaking at an arbitrary character count is simpler and in fact affords the opportunity for a single chunk containing an extended argument or discussion. 
The overlapping of chunks guards against the error of mid-sentence truncation, and increases the probability that a sentence relevant to the user query is accompanied by the required contextual information. This overlapping means there is some redundancy between chunks, but this is a small inefficiency. 
Each chunk is converted to an embedding vector by a lookup in a text embedding model. The set of chunks and corresponding vectors are stored in a database for later retrieval.

\begin{figure*}
\centering
  \includegraphics[width=10.0cm]{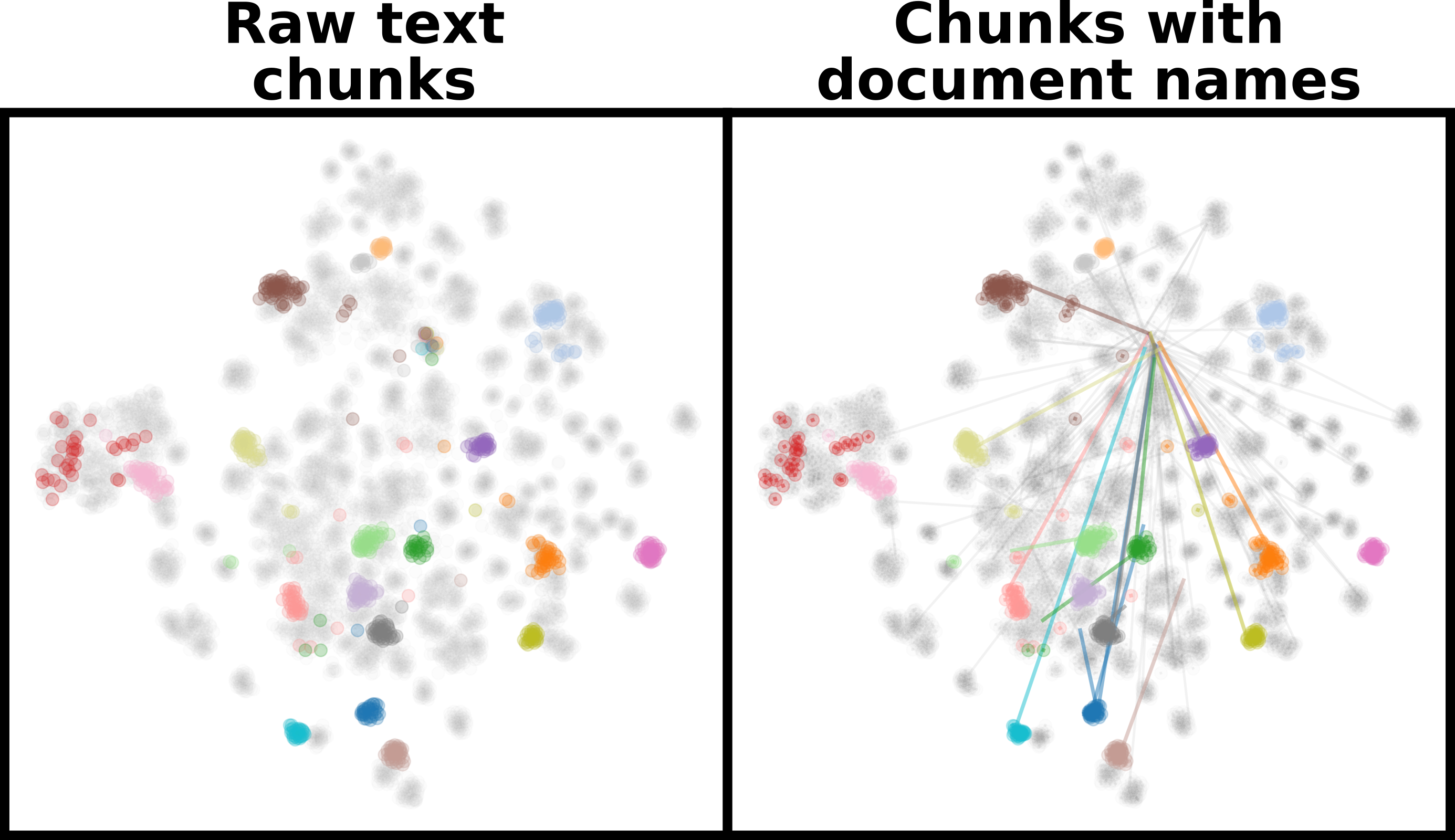}
  \caption{(left) The text chunks derived from documents can be positioned in a high-dimensional ($1,536$) semantic space. For visualization, this space is projected into two dimensions using the t-SNE method.\cite{JMLR:v9:vandermaaten08a} Each chunk is visualized using a grey dot, but twenty (randomly selected) documents are assigned a particular color. The grouping of chunks from a particular document confirms that the text embedding is succeeding in capturing meaning. A small number of chunks are far from their document cluster. (right) By prepending the document name to each text chunk before computing the embedding, otherwise ``orphaned'' chunks are grouped with other chunks from that document. The displacement of chunks from their raw position to this improved position is shown using a connecting line. While most chunks are not displaced, orphaned chunks become correctly grouped.
}
  \label{fgr:names}
\end{figure*}

A valid concern with such a procedure is that many chunks will be ``orphaned'' in that their content will lack context and be correspondingly meaningless when read alone. Such chunks might contain useful information, but would not be reliably retrieved since their isolated content would not be semantically similar to the user query. 
A simple improvement to naive chunking is to prepend to each chunk the document name, and use that augmented chunk for embedding calculation and later retrieval. This anchors each chunk to the context provided by the title, and allows the eventual chatbot LLM to identify the source of each provided chunk. 
In Figure~\ref{fgr:names}, we visualize the distribution of text chunks in the embedding space. Since the high-dimensional ($1,536$) embedding space cannot be easily understood, we project it into a two-dimensional (2D) space using the t-distributed stochastic neighbor embedding (t-SNE) method,\cite{JMLR:v9:vandermaaten08a} which stochastically redistributes points while maintaining pairwise similarities and thus clustering. 
What can be observed (Fig.~\ref{fgr:names}, left) is that the different documents are naturally clustered, confirming that the embedding space provides a semantically meaningful organization. However, a small number of document chunks are found to be extremely far from the centroid cluster for their parent document. 
When recomputing the embeddings with the document name, these orphaned chunks are typically returned to the same neighborhood as the other chunks from that document, confirming that this strategy is helpful in maintaining context for each chunk.

Overall, this chatbot configuration provides a robust means of delivering meaningful answers to user queries. When comparing raw LLM output to the LLM with context chunks (refer to ESI section 1 for examples), we find a vast improvement in the quality of responses by providing chunks. 
The raw LLM is prone to fabricating plausible-sounding answers that are nonsense (including adding citations to nonexistent papers), making the responses unsuitable for serious scientific research. 
However, with access to contextual information, the LLM much more reliably provides true answers (drawn from the provided text) and more helpful analysis. 
The quality of the chatbot response thereby becomes limited by the quality of the provided context chunks, which is limited by the size of the context window, and the quality of the embedding similarity lookup. In general, we find that embedding lookup is successful in identifying relevant document extracts; although it is not guaranteed to find all relevant chunks, especially for complex queries. 
Thus, there are clearly opportunities to refine these methods by more carefully selecting and aggregating contextual information into the prompt.

\subsection{Prompt engineering}

\begin{figure*}
\centering
  \includegraphics[width=10.0cm]{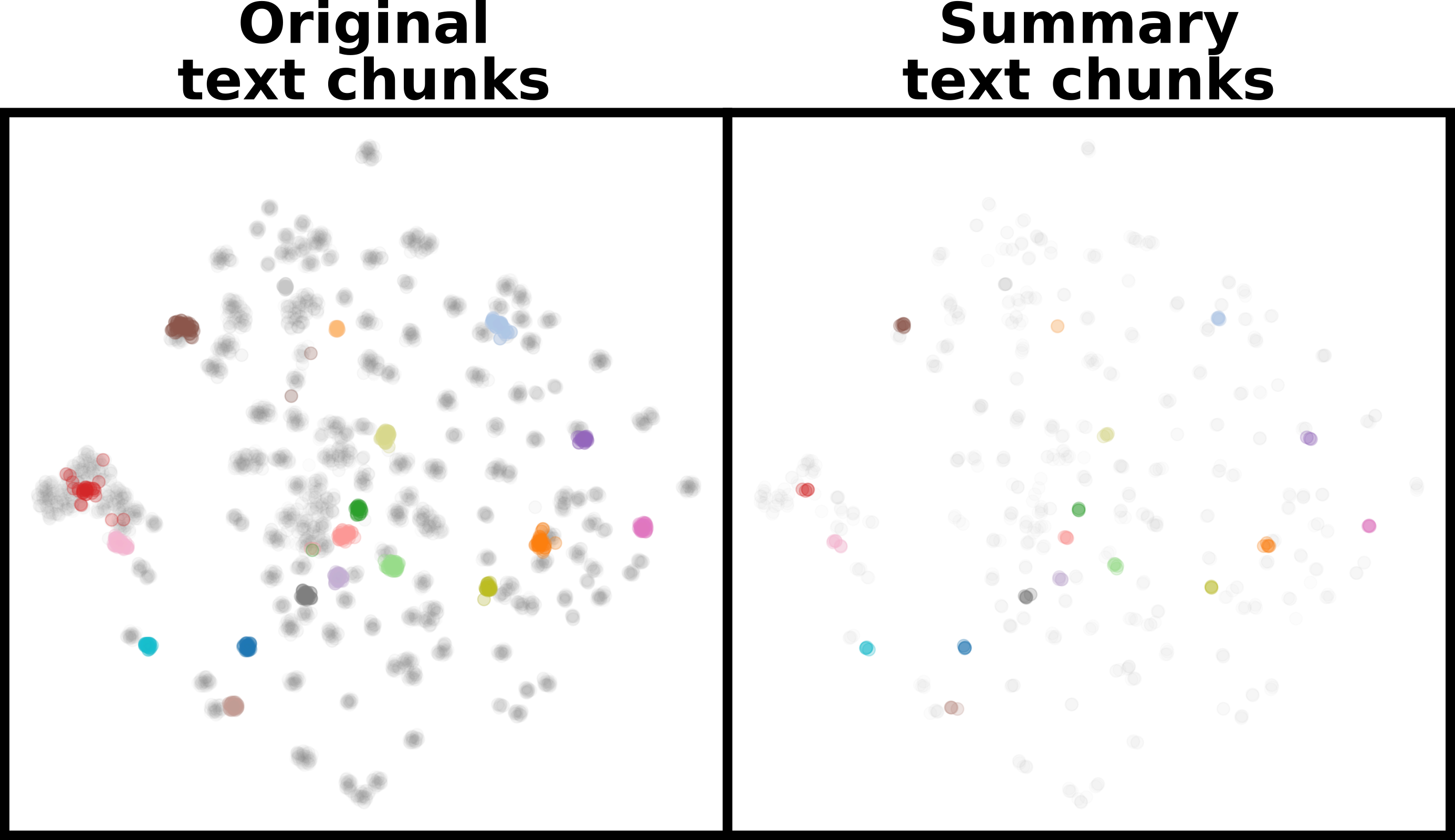}
  \caption{(left) The $6,157$ text chunks obtained from $176$ documents are clustered in a semantic space (2D t-SNE projection shown, with 20 documents randomly assigned colors). (right) Rather than generating chunks from the raw document text, a set of $707$ chunks can be created from LLM summaries of the original documents. This much smaller number of chunks are organized identically in the semantic space, suggesting that the meaning of the original documents is being preserved during summarization.
}
  \label{fgr:summaries}
\end{figure*}

Instead of using the raw document text, other strategies can be pursued. 
Scientific documents often contain some informational redundancy, e.g. due to writing style. 
For instance, careful and elaborated arguments may be used in documents to educate readers or guide them to a conclusion; whereas only a concise summary of key findings might be relevant for tasks such as giving a chatbot contextual data. 
Thus, one option in building the chunk database is to create a store of more concise or ``compressed'' text chunks, which should correspondingly allow more concepts to be placed in the LLM context window. 
We investigated this possibility by generating a set of summary chunks, by passing each text extract to an LLM with instructions to summarize the chunk (Fig.~\ref{fgr:chunking}b). 
In terms of distribution in the embedding space, we find that this summarization step retains semantic meaning (Figure~\ref{fgr:summaries}), at least at a coarse level, while greatly reducing the size of the corpus (by a factor of $\approx 10 \times$).

Although this summarization procedure is an attractive option for increasing lookup speed (smaller corpus size) and response quality (more chunks in prompt), we find that in general the chatbot response quality suffers (refer to ESI section 2 for examples). 
With access to only summarized information, the chatbot occasionally makes small errors---essentially misinterpreting results or misunderstanding their context---owing to the successive text reinterpretation. In this sense, access to raw text material is preferable. 
Interestingly, we observe that providing access to both raw and summarized information is useful. While this procedure means the prompt ends up with redundant information, this may serve to reemphasize salient points (phrased in different ways) and guard against mistakes. 
There is growing evidence that LLMs effectively perform analysis within the output generation itself.\cite{kojima2023large} Thus, LLM prompts that suggest (for instance) ``Let's think step by step'' increase respone quality by inducing the LLM to build a chain of reasoning in the output. Similarly, providing pre-computed LLM rewording of text chunks affords the system an opportunity to pre-build some textual analysis. 

There are ample opportunities for further increasing the performance of domain-specific chatbots by more carefully crafting the chunking and prompt construction steps. For instance, rather than simple summarization, an LLM could be used to preprocess input documents in more sophisticated ways, such as specifically extracting items of interest, or performing contrastive analysis between chunks from different (but related) documents. Or, the embedding of the user query could be used to select among different construction strategies (or even among different LLM systems for the generation step). 
These and other refinements will no doubt be studied in the near future, thereby refining chatbot performance. However, even the simplest implementation presented here is already able to reliably identify useful documents (and sub-sections thereof) and converse about that content in a useful manner.

\subsection{Usage Evaluation}

It is worth briefly considering an alternate strategy to the embedding lookup described here. In particular, one could imagine inputting the entire text corpus into the LLM, allowing it to select relevant text through the transformer attentional mechanism. 
A crucial limit to LLMs is the finite context window, into which one must add completion instructions and relevant context data. 
Typically available LLMs have context windows of 4k tokens to 100k tokens (where a token is the atomic unit of LLM parsing; typically a word or word-fragment). 
Such a context window is insufficient for a corpus of multiple scientific publications (the test corpus used here of 176 documents is >1M words in length). 
There is exciting ongoing research into greatly extending the context window of LLMs.\cite{beltagy2020longformer, ding2023longnet, tworkowski2023focused} However, even with larger context windows, there may remain advantages to using embedding strategies to isolate relevant text. Firstly, larger context window sizes increase computational cost and completion time; embedding lookup is typically faster since embeddings are precomputed. Secondly, there are open questions about how the attentional mechanism behaves in extremely large context windows; whereas embedding lookup allows the user to craft retrieval to their particular needs. 
Nevertheless, it is clear that expanded context windows will yield enormous benefits for domain-specific chatbots, by allowing them to reason about larger and more complex sets of text data; e.g. performing more sophisticated comparison between publications, or summarizing an entire domain of research.

ChatBots and LLM systems more generally operate on natural language data, and provide complex linguistic replies that are typically non-deterministic. This makes rigorous evaluation difficult. Nevertheless, many efforts have been formulated to quantify LLM performance,\cite{clark2018think, zellers2019hellaswag, hendrycks2021measuring, lin2022truthfulqa} and enable ranking of implementations. 
Rigorous evaluation of science-specific usage is challenging owing to the lack of available community testing datasets. To assess the advantages of the present implementation, the test questions and answers would need to be tailored to the input document corpus, since questions outside this domain the system would simply revert to the general capabilities of the underlying language model. 
In our testing, we found that using our optimal embedding strategy and reasonable queries (that a human could answer by looking through the documents), the system usually returns responses that are valid and lack hallucinations ($\approx 90\%$ success, as compared to $<14\%$ success without embedding).

To further evaluate the LLM, we devised other quantitative tests. Language models can be used to sort documents by arbitrary and imprecise criteria. An efficient and scalable strategy is to repeatedly ask the LLM to perform pairwise comparisons, and use this set of comparisons to construct an ordering.\cite{qin2023large} The set of comparisons need not be exhaustive, and additional documents can be added to the list with only a small number of additional comparisons (to identify the location in the sorted list for the new item). 
We tested the ability of an LLM to sort scientific publications by predicted impact; we find that its output roughly correlates to the impact factor of the journal the work was published in, implying that the LLM is capturing some aspects that humans use to predict impact (ESI Figs~1 and 2).

\begin{table*}[h]
\centering
{\small
\setlength{\tabcolsep}{3pt}
\begin{tabular}{c|cccccc|ccc}
\hline
\textbf{Ground truth} & \multicolumn{6}{c|}{ \textbf{LLM assignment} } & \multicolumn{3}{c}{\textbf{metrics}} \\
\hline
 & \small{Self-assembly} & \small{Materials} & \small{Scattering} & \small{Machine-learning} & \small{Photo-responsive} & \small{Other} & \textit{Pr} & \textit{Re} & \textit{Ac} \\
\hline
 Self-assembly & \cellcolor{goodgreen} \textbf{95\%} & 3\% & \textcolor{lightgrey}{0\%} & \textcolor{lightgrey}{0\%} & \textcolor{lightgrey}{0\%} & 2\% & 79\% & 95\% & 89\% \\
\hline
 Materials & 28\% & \cellcolor{goodgreen} \textbf{53\%} & 17\% & 2\% & \textcolor{lightgrey}{0\%} & \textcolor{lightgrey}{0\%} & 84\% & 53\% & 81\% \\
\hline
 Scattering & \textcolor{lightgrey}{0\%} & \textcolor{lightgrey}{0\%} & \cellcolor{goodgreen} \textbf{100\%} & \textcolor{lightgrey}{0\%} & \textcolor{lightgrey}{0\%} & \textcolor{lightgrey}{0\%} & 41\% & 100\% & 91\% \\
\hline
 Machine-learning & \textcolor{lightgrey}{0\%} & 4\% & 22\% & \cellcolor{goodgreen} \textbf{70\%} & \textcolor{lightgrey}{0\%} & 4\% & 94\% & 70\% & 95\% \\
\hline
 Photo-responsive & \textcolor{lightgrey}{0\%} & 9\% & \textcolor{lightgrey}{0\%} & \textcolor{lightgrey}{0\%} & \cellcolor{goodgreen} \textbf{91\%} & \textcolor{lightgrey}{0\%} & 100\% & 91\% & 99\% \\
\hline
 Other & \textcolor{lightgrey}{0\%} & 40\% & 20\% & \textcolor{lightgrey}{0\%} & \textcolor{lightgrey}{0\%} & \cellcolor{goodgreen} \textbf{40\%} & 50\% & 40\% & 97\% \\
\hline
\end{tabular}
}
\caption{ Evaluation of the ability of LLM (OpenAI GPT 3.5) to classify scientific documents. Each document was manually classified into one of 6 thematic categories. The central $6\times6$ cells show the distribution of LLM classifications. The rightmost columns provide prediction metrics, including precision (\textit{Pr}), recall (\textit{Re}), and accuracy (\textit{Ac}). Overall, the LLM is successful at this imprecise task. }
  \label{tbl:class}
\end{table*}

As another test, the LLM was tasked with assigning the scientific documents into a set of human-selected categories. This classification task can be compared to human selections for the same task, in order to quantify performance (Table \ref{tbl:class}). This is an inherently imprecise task, especially given the overlap in the selected categories. Nevertheless, the LLM is highly successful at this challenging task (accuracy 81-99\%), with the majority of errors being reasonable (e.g. ambiguous classification between \textit{materials} category, or \textit{self-assembly} category more specifically).
The strong performance across a diverse set of tasks, as presented here, helps to support the argument that a chatbot with access to domain-specific documents can assist researchers in a variety of meaningful tasks.

\subsection{Image data}

\begin{figure*}
\centering
  \includegraphics[width=12.0cm]{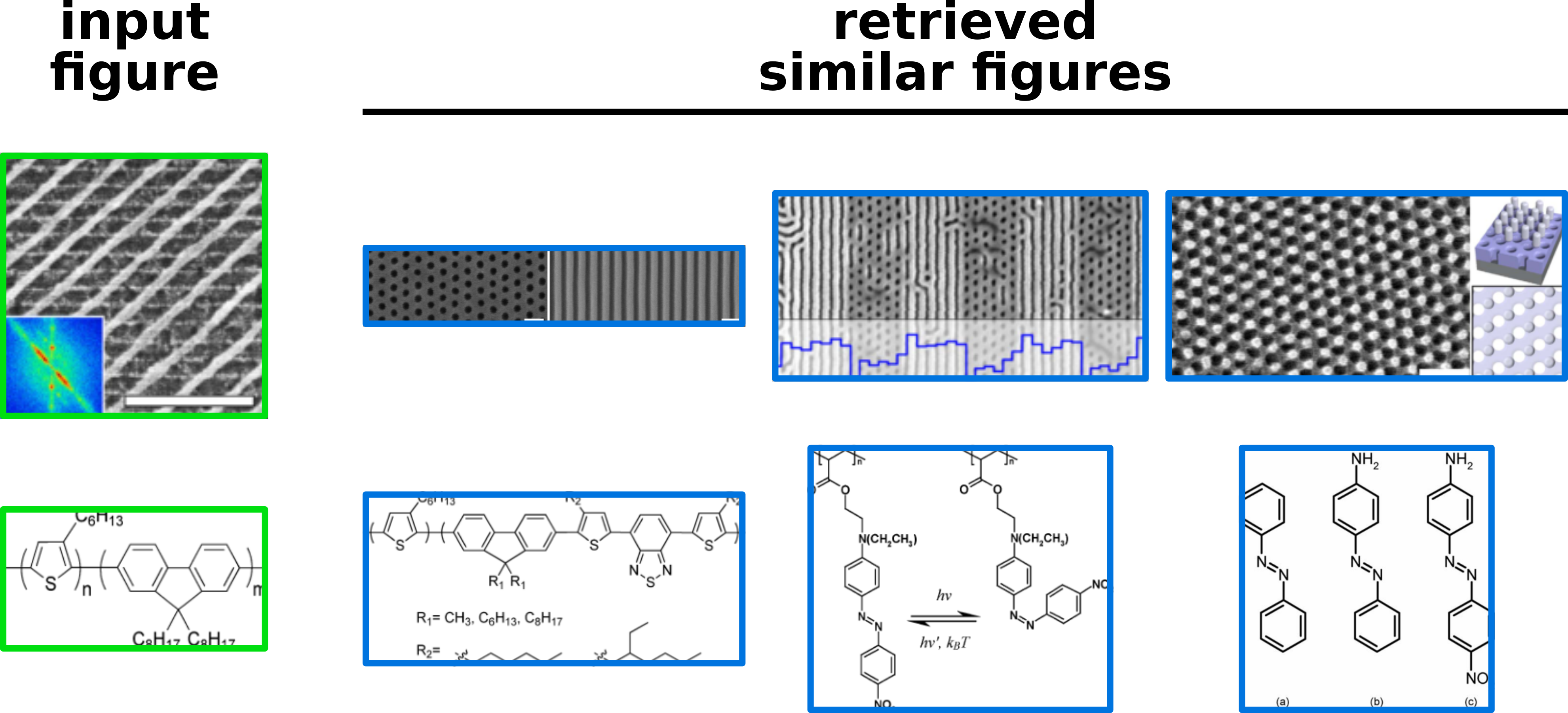}
  \caption{Image embeddings can be used to identify semantically related figures (or portions thereof) from a datastore of pre-processed documents. An electron micrograph of a mesh nanostructure (Fig. 3d from \cite{Majewski2015mesh}) yields suggestions of other micrographs of nanostructure arrays (Fig. 3h-i from \cite{Stein2016}, Fig. 5b from \cite{Stein2016}, Fig. 3e from \cite{Rahman2016nonnative}). The chemical structure of a polymer (Scheme 1 from \cite{C3SM53090F}) yields chemical structures from other publications in the database (Fig. 1 from \cite{Smith2015MolecularOrigin}, Fig. 1 from \cite{Yager2006macromol}, Fig. 1 from \cite{YAGER2006250}). Cosine similarity was used to identify relevant images.
}
  \label{fgr:figures}
\end{figure*}

\begin{figure*}
\centering
  \includegraphics[width=12.0cm]{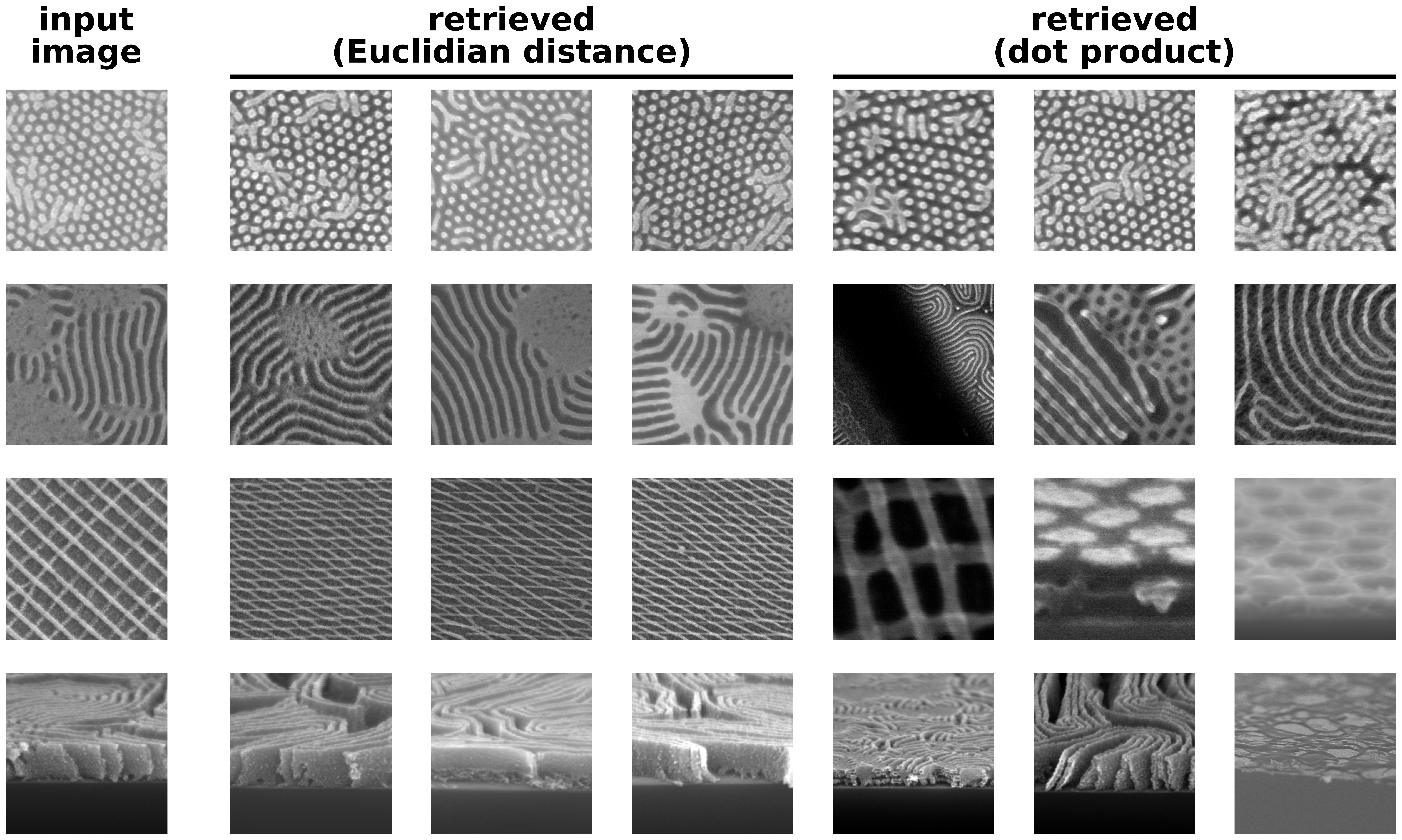}
  \caption{Image embedding and retrieval can be applied to arbitrary image data. Scanning electron microscope (SEM) micrographs for samples were used as image inputs, to search for semantically similar images in a pre-computed dataset of $20,302$ other SEM images. The retrieval is rapid and meaningful, with relevant figures being retrieved. Crops of the SEMs are shown for clarity; the embeddings were computing on the full SEM image. Examples are shown for retrieval using Euclidian distance (which measures similarity in assessed meaning), as well as dot product similarity (measures overlap in concepts).
}
  \label{fgr:images}
\end{figure*}

A key aspect of research, and scientific publications, is the data visualization used to reason about trends and describe results to others. Thus, it is important to consider how to aid researchers in image-based search and retrieval. 
Image embedding methods can generate a vector describing an image's semantic content, similar to text embedding, allowing retrieval of related images or other image-image reasoning operations. These capabilities can be exploited to help researchers perform search and retrieval related to the figures contained in scientific documents.
To demonstrate the utility of image embedding for scientific documents, we added a simple image similarity search component to our system. 
The publication figures identified by Grobid were extracted from the input PDF documents, and converted into an image embedding vector using the Contrastive Language-Image Pre-Training (CLIP)\cite{radford2021learning} method. 
Because CLIP is a multi-modal embedding trained on images and text, it acquires visual understanding embodying meaningful semantics. This affords the opportunity for similarity measures based on human concepts. 
This approach enables the user to input an image, and find publication figures with semantic similarity (Figure~\ref{fgr:figures}). 
Importantly, these search and retrieval operations are performed on the user-provided corpus of scientific documents, allowing tasks to be much more domain-specific than when using generic (e.g. web-based) image search systems.
In addition to exploring publication figures, one can compute image embeddings for a set of unlabelled experimental data (such as electron micrographs, scanning probe images, or scattering/diffraction detector images) and thereby perform similarity lookup on input images (Figure~\ref{fgr:images} and ESI Figs~3 and 4). 
This method is extremely successful in rapidly identifying relevant images, capturing aspects of similarity well beyond simple pixelwise or sub-structural match. 
Moreover, the user can select among similarity measures to achieve different kinds of retrieval. For instance, Euclidian distance in the embedding space assesses how similar images are, while the dot product between embedding vectors measures a looser kind of overlap between the underlying concepts. 
This retrieval can be viewed as a form of zero-shot ML, in the sense that the CLIP model was not trained on scientific images explicitly, and yet it can provide a meaningful descriptor of these images. In other words, the semantic understanding of CLIP is sufficiently broad and robust that it generalizes to the kinds of images used in scientific contexts. 
This suggests that these existing models can be immediately deployed on scientific instruments to assist in organizing and classifying images, and can be valuable for searching through publications to find relevant data (e.g. to find examples of a particular kind of measurement result).

\section{Perspectives}

We have demonstrated that existing technologies are already suitable for use by physical scientists to speed up their work in identifying useful publications (and sub-sections or figures therein), and to provide rapid and meaningful answers to scientific questions. 
By using a domain-specific chatbot, with access to relevant documents, scientists should be able to accelerate several aspects of their workflow. For instance, such a digital assistant should be useful during literature searching, proposal writing, manuscript drafting, hypothesis testing, and ideation.\cite{haase2023artificial, girotra2023ideas, boussioux2023crowdless, doshi2023generative}

The presented implementation is simple and quite limited; it could be improved in several ways.
We have used a traditional relational database to store text and embeddings. 
For the modest corpus sizes considered here, lookup time is not performance limiting. Yet a more scaleable solution would be to use a special-purpose vector databases (such as Pinecone,\cite{PineconeURL} Milvus,\cite{MilvusURL} or Chroma\cite{ChromaURL}).
As previously discussed, there are opportunities to further refine the presented prompt construction strategy. 
Herein, we have used a cloud LLM; an alternative would be to deploy an LLM locally to avoid the latency and privacy concerns associated with cloud lookup.\cite{vicuna2023, localGPTURL} More targeted chatbot behavior could also be obtained by fine-tuning a local LLM by retraining on the relevant corpus of documents. Note that even with fine-tuning the LLM itself, there remains an advantage in providing text extracts in the constructed prompt. Namely, it provides relevant information to assist LLM reasonining, and enables direct quoting and citation. 
It is also obvious that as LLMs increase in sophistication (as previously discussed), the corresponding domain-specific chatbots will correspondingly increase in sophistication. This suggests that domain-specific endeavors should in fact be designed in a way that is decoupled from any specific LLM implementation. This will allow them to take advantage of future improvements in LLMs by simply changing the system being accessed.

An exciting possibility afforded by LLMs is literature-based discovery (LBD),\cite{wang2023learning} which seeks to make discoveries by mining the literature and proposing/testing hypotheses. 
For instance, trends or commonalities can be automatically extracted from the literature by LLM scanning the literature, generating lists of conclusions, and aggregating the results. 
A chatbot can acceleration human ideation, by providing immediate feedback on hypotheses, and retrieving relevant documents from the literature. 
We expect to see increasingly sophisticated literature discovery paradigms emerge as domain-specific chatbots are deployed more broadly.

An emerging trend in experimental sciences is autonomous experimentation (AE), wherein the measurement loop is closed using a decision-making algorithm that selects high-quality experiments to perform.\cite{stein_progress, STACH20212702, Abolhasani2023} 
For instance, researchers have demonstrated that a synchrotron x-ray scattering beamline can autonomously explore physical parameter spaces,\cite{noack_autonomous_2020, Yager_2023} reconstructing a high-quality model of the space and even discovering new materials or structures.\cite{Doerk2022AE} 
Existing approaches have typically used grounded ML modeling approaches (such as Gaussian process regression). It is interesting to consider whether the more flexible and general-purpose understanding of LLMs can be directly leverage as a decision-making agent in experimental loops. There is early evidence that LLMs can indeed engage in autonomous scientific discovery,\cite{boiko2023emergent} and further elaboration of these methods is an exciting avenue for future study.

\section{Methods}

\subsection{Database preparation}

For testing, we assembled a dataset of 176 PDF files (the author's full set of peer-reviewed scientific journal publications and book chapters), with a cumulative file size of 703 MB.  
We use the open-source Grobid\cite{GROBID} system to convert these PDF files into XML files. The XML files were parsed using the Python library BeautifulSoup. 
For further analysis, only the main text was considered, eliminating input PDF boilerplate and references sections. In total, the text corpus is $1,061,967$ words ($\approx 3,500$ pages of textual data). 
Text chunks were generated by breaking the input document main text into segments $1,400$ characters in length, with an overlap of $280$ characters between subsequent chunks. The overlap accounts for the random truncation of sentences, and increases the probability of a given block of text being found in a chunk along with relevant contextual information. 
An embedding vector was computed for each text chunk using the OpenAI cloud API, and the \texttt{text-embedding-ada-002} model, which returns a $1,536$ length vector. The text chunks and vectors were stored in a MySQL database. 
For retrieval efficiency, the list of embedding vectors was cached in a binary file using the numpy Python library.\cite{Oliphant4160250}

Summaries of raw text chunks were obtained by calling the \texttt{gpt-3.5-turbo} model (OpenAI) with a prompt that included instructions to "summarize in a concise way." These summaries were concatenated into a summary document, which was in turn chunked. Embeddings for each chunk were computed as before. Thus, the chunk summaries are smaller in number than the raw text chunks, representing a substantial compression of the original text ($707/6,157 \approx 11 \%$).

\subsection{Visualization}
Visualization of the semantic organization of document chunks was performed using the t-distributed stochastic neighbor embedding (t-SNE) method.\cite{Hinton_tSNE, JMLR:v9:vandermaaten08a} This computes a statistical non-linear mapping of points from a high-dimensional space into a lower-dimensional space, attempting to maintain pairwise similarity. We project from the $1,536$ dimensional text embedding space defined by the embedding model into a two-dimensional (2D) space, using a perplexity of 40 and $10,000$ iterations. Images were plotted using the matplotlib\cite{Hunter4160265} package.

\subsection{Chatbot querying}

The results discussed primarily used the \texttt{gpt-3.5-turbo-0301}, accessed via the OpenAI cloud API using Python code. 
We assume an overall context window of 16,384 characters (based on a model limit of 4,096 tokens). 
Prompts were constructed by providing an instruction to answer user queries using provided text, followed by a sequence of text extracts, followed by the user query. Relevant chunks were identified based on cosine similarity (Eq.~\ref{eq:cosine}). By assessing the relative angle between vectors, this measure assesses thematic similarity. Since the selected embedding is normalized, this is equivalent to Euclidian distance. As many chunks as possible were added to the context window, while reserving $3,564$ characters ($\approx 900$ tokens) for chatbot response. A single query executes in $\approx 10 \, \mathrm{s}$, with the chunk lookup and prompt construction requiring $< 1 \, \mathrm{s}$, and the majority of execution time resulting from the cloud LLM lookup.

For comparisons of response quality (refer to ESI), we also tested the ChatGPT4 model, using the web interface. In this case, constructed prompts were manually copied into the web interface (using a new conversation thread with no history) to generate a response. The ChatGPT4 system uses the more powerful and knowledgeable GPT4 model, and also leverages the fine-tuning performed by OpenAI in developing the ChatGPT versions of their system. This system emphasizes useful responses, while minimizing fabrications.

The models used herein have a temperature parameter that can be used to influence model output (refer to ESI for examples). Low values of this parameter have limited variability and induce more deterministic output. Higher values of this parameter lead to more variable output, with sufficiently large values leading to outputs corrupted by irrelevant text completions. The default value (1.0) was used for the presented results (except where noted otherwise), and was found to yield reasonable responses for the tasks explored. Repeatability tests confirm that at this setting, the model output varies in exact wording, but retains the same general semantic meaning.

\subsection{Evaluation}
Input questions were manually crafted and selected in an attempt to cover a distribution of use-cases relevant to the training corpus. Questions were constructed such that a science-trained human would be able to answer them if they were familiar with the input documents and given time to look through the documents, but without spending time performing extensive background research or thinking. 
Model outputs were manually scored to identify components that were incorrect (or fabricated) versus correct. Answers were judged overall correct when they provided valid information without introducing erroneous ideas. Based on this manual assessment, it was found that the embedding strategy can respond successfully to $\approx 90 \%$ of queries, which can be compared to a $< 14 \%$ success rate when embedding is not performed.

In order to sort documents by scientific impact, a set of pairwise comparisons were generated, where for each comparison the LLM is asked to select which publication is higher impact. 
The LLM was provided with each document's text---including title, abstract, and initial portion of main text (up to the context limit of the model)---but not provided with ancillary information such as journal name. 
Comparison pairs were selected randomly, biased so that every document is involved in at least one pairwise comparison. From this set of 818 comparisons (out of a total possible $176^2=30,976$), a ranked list of documents was generated through a straightforward sorting procedure; namely, iteratively considering pairs of documents, and swapping their order if the swap reduces (or does not change) the total number of misordered pairs (i.e. pairs where a higher-impact paper is incorrectly sorted lower in the list). This procedure does not resolve to zero misordered pairs, since the LLM pairwise comparisons are not guaranteed to form a perfectly consistent set. Viewed as a directed graph, we indeed identify cycles. 
Nevertheless, the ordering is found to be meaningful, as it roughly correlates to the impact factor of the journal the work was published in, implying that the LLM is capturing some aspects that humans use to predict impact (ESI Figs~1 and 2).

The LLM was evaluated on a classification task, where each document was manually assigned to one of 6 thematic categories. The LLM was then asked to classify each document into one of those categories. The task is inherently ambiguous, since some categories are subsets of others (e.g. \textit{self-assembly} and \textit{photo-responsive} papers are special cases of the more general \textit{materials} category), while other publications touch on multiple topics (e.g. some papers involve applying \textit{machine-learning} to \textit{scattering} datasets). 
Despite this challenge, the LLM identifies the same category as the human in the majority of cases (accuracy 81-99\%).

\subsection{Image querying}
Figures from publications were identified from the Grobid XML files, and extracted from the PDF documents into images using the PyMuPDF library. References to figures, along with captions, were stored in the MySQL database. Raw images were similarly added to the database, without caption information. Image embeddings were computed using the Contrastive Language-Image Pre-Training (CLIP)\cite{radford2021learning} method, specifically the \texttt{ViT-B/32} pre-trained model provided by the PyTorch\cite{pytorch} deep learning environment (vectors are length $512$). Bulk calculation of embeddings requires $\approx 66 \, \mathrm{ms}$ per image. The list of embeddings were stored in the MySQL database. Image similarity was computed using multiple measures: Euclidian distance in the CLIP space, which measures the distance between the meaning of the images; cosine similarity of the embedding vectors, which measures the thematic similarity; and the raw (non-normalized) dot product, which measures a form of projected relatedness. Since CLIP embeddings are not normalized, the cosine similarity and Euclidian distances are not equivalent; nevertheless in practice they are found to return highly similar results, since the image corpus is relatively clustered in the overall CLIP space.

\section*{Code Availability}
Source code for chatbot and associated tools is available at https://github.com/CFN-softbio/SciBot.

\section*{Author Contributions}
KGY developed the concepts, authored the software, conducted the research, and wrote the manuscript. A LLM chatbot (ChatGPT4) was used as a digital assistant to expedite code development.

\section*{Conflicts of interest}
There are no conflicts to declare.

\section*{Acknowledgements}
This research was carried out by the Center for Functional Nanomaterials, which is a U.S. DOE Office of Science Facility, at Brookhaven National Laboratory under Contract No. DE-SC0012704. We thank Dr. Kevin F. Casey for fruitful discussion. We acknowledge the important contributions of the open source software community.

\bibliography{rsc} 
\bibliographystyle{rsc} 

\end{document}


{\centering
\null \vspace{3.5 cm}
\noindent \textbf{\huge Supplementary Information for: Domain-specific ChatBots for Science using Embeddings}
\par \null \vspace{1.5 cm}
\noindent \textbf{\Large }
\par \null \vspace{1.5 cm}
\noindent {\Large Kevin G. Yager$^{\ast}$}
\par \null \vspace{0.5 cm}
\noindent {\large Center for Functional Nanomaterials, Brookhaven National Laboratory, Upton, New York 11973, United States}
\par\null \vspace{0.5 cm}
\par\null\par
}
\clearpage
\par

\section{Examples of Chatbot Performance}

Below we provide examples of chatbot responses to various scientific questions. We compare unaided responses (i.e. without providing any text chunks from the document store) to contextualized prompts (i.e. where text chunks are added to the prompt). We also compare the GPT-3.5 model (accessed via the OpenAI API) and ChatGPT-4 (accessed via the web interface). The LLM responses are unedited, other than reformatting to convert symbols into \LaTeX. Highlights were added to the response text. \hlg{Green highlights} show correct responses highly-matched to the user query. \hlb{Red text} highlights erroneous parts of the response (highly mismatched to user needs or delivering incorrect information). Reasonable and factually correct (though not always optimal or necessary) parts of the response are left unhighlighted.

\subsection{Example: Answer scientific question}
\textbf{User Query:} How can I use x-ray scattering to measure the grain size?
\begin{itemize}
    \item \textbf{GPT-3.5 (unaided) response:}
    
X-ray scattering can be used to measure the grain size of a material by analyzing the diffraction pattern produced when a beam of X-rays interacts with the material's crystalline lattice structure. As the X-rays pass through the material, they are diffracted by the crystal planes and produce a pattern of peaks that can be analyzed to determine the crystal structure and grain size.

\hlb{According to a paper published in Materials Science and Engineering: R: Reports}, " The grain size can be calculated using the Scherrer equation, which \hlg{relates the peak width to the grain size}, or by more advanced methods such as the Warren-Averbach method or the Williamson-Hall method." [1]

There are also several specialized X-ray diffraction techniques, such as small-angle X-ray scattering (SAXS) and grazing-incidence small-angle X-ray scattering (GISAXS), that can provide additional information on the size and shape of nanocrystals and thin films. 

\hlb{Reference:
[1] Salgado-Morán, G. et al. (2017). In situ and Operando X-ray Techniques in Nanomedicine: From Preclinical Development to Advanced Clinical Applications. Materials Science and Engineering: R: Reports, 114, 1-46. https://doi.org/10.1016/j.mser.2016.10.001}

\textbf{Analysis}: Fabricated (``hallucinated'') a non-existent reference. Did not directly explain the method/calculation. \textbf{Poor response}.

    \item \textbf{GPT-3.5 (with context chunks) response:}

X-ray scattering is a powerful technique that can be used to measure structural order at the atomic/molecular scale through wide-angle X-ray scattering (WAXS) and at the nanoscale through small-angle X-ray scattering (SAXS). X-ray scattering can be used to measure the average structure throughout a material, making it complementary to real-space imaging techniques, which may provide only local measurements difficult to generalize. 

To measure grain size using X-ray scattering, one can use conventional X-ray powder diffraction measurements, which yield distinct rings on the detector. The ring position, $q = (4\pi/\lambda)\mathrm{sin}\theta$, provides information about the spacing of structures ($d = 2\pi/q$). \hlg{The average grain size can be estimated from the peak width.} One method is the Debye-Scherrer analysis, which allows extraction of a coherence length from the peak width using \hlb{$\lambda/\Delta\lambda \approx 2K/q$}, where the Scherrer constant $K$ is of order unity. 

There is a peak width limit for the Scherrer analysis, however, meaning that the grain size that can be measured with this method is typically limited to the submicron range for modern synchrotron beamlines. An alternative method called \hleg{"variance scattering" has been proposed, which quantifies the statistical variation of intensity along a scattering ring}. This method enables quantification of grain size across a wide range of sizes, from nanometers to hundreds of microns, making it useful for materials with extremely large or micrometer-sized grains.

\textbf{Analysis}: Explained method using a partially incorrect equation. Provided information about alternate methods from source documents. \textbf{Partial success}.

    \item \textbf{ChatGPT-4 (unaided) response:}

X-ray scattering techniques, such as X-ray diffraction (XRD), can be used to measure the grain size in a polycrystalline material. This is often achieved through the application of \hlg{the Scherrer Equation, which relates the size of the crystallites in a sample to the broadening of the diffraction peak}.

The key steps in this process include:

1. \textbf{X-Ray Diffraction Experiment}: The first step is to conduct an XRD experiment on your sample. The sample is bombarded with X-rays and the diffraction pattern is recorded.

2. \textbf{Peak Identification}: Identify the diffraction peaks in the XRD pattern. These peaks correspond to different planes of atoms in the crystalline material, and the peak broadening is related to the grain size.

3. \textbf{Measure Peak Broadening}: The width of the diffraction peak at half its maximum height, known as the Full Width at Half Maximum (FWHM), is measured. This is typically done in radians. 

4. \textbf{Application of Scherrer Equation}: The Scherrer Equation is typically written as:

    \hlg{$\tau = K\lambda / (\beta\mathrm{cos}\theta)$}

    Where:
\begin{itemize}
    \item $\tau$ is the crystallite size
    \item $K$ is the shape factor, with a usual value of about 0.9, but it depends on the specific crystallite geometry
    \item $\lambda$ is the X-ray wavelength
    \item $\beta$ is the line broadening at half the maximum intensity (FWHM), after subtracting the instrumental line broadening, in radians
    \item $\theta$ is the Bragg angle (the angle between the incident ray and the scattering planes)
\end{itemize}
Note that \hlg{the Scherrer Equation assumes that the peak broadening is only due to the crystallite size, but in reality, there could be other factors contributing to the broadening, such as strain within the material.} So, the grain size calculated using this method should be taken as an approximation. More accurate methods would involve additional corrections or use more advanced techniques such as electron backscatter diffraction (EBSD).

Also remember that this method provides the average size of crystallites that are contributing to the diffraction peak, not necessarily the average grain size of the entire sample, as not all grains contribute equally to the XRD pattern.

\textbf{Analysis}: Detailed and accurate answer, using general knowledge. Provided additional information and caveats. \textbf{Good response}.

    \item \textbf{ChatGPT-4 (with context chunks) response:}

X-ray scattering is a powerful technique to measure grain size in various materials. The fundamental concept is that when X-rays are incident on a crystalline material, they are diffracted in different directions based on the periodic arrangement of atoms. This diffraction pattern contains information about the size of the grains within the material.

There are two primary methods of grain size estimation using X-ray scattering mentioned in the sources you provided:

\textbf{1. Scherrer analysis}: This method uses the peak width of the diffracted X-ray pattern. The broadening of these peaks provides information about the average size of the crystalline domains or 'grains' in the material. The Debye-Scherrer equation (\hlg{$D = K\lambda / \beta \mathrm{cos} \theta$}) can be used, where $D$ is the crystallite size, $K$ is the shape factor (Scherrer constant), $\lambda$ is the X-ray wavelength, $\beta$ is the full width at half maximum (FWHM) of the peak, and $\theta$ is the Bragg angle. \hleg{However, this method has limitations and can usually only measure submicron grain sizes, as instrumental effects such as beam divergence impose a minimum peak width.}

\textbf{2. Graininess or non-uniformity method}: This newer method, discussed by Yager, Majewski, et al., relies on quantifying the non-uniformity of X-ray scattering rings, which arise from the finite number of grains in the probed volume. In essence, if you have many small, randomly-oriented grains, they give rise to a uniform scattering ring as the numerous Bragg peaks overlap. In contrast, if you have a smaller number of larger grains, you get a scattering ring with significant intensity variation. \hleg{By looking at this variance or non-uniformity in the scattering ring, you can estimate the number of grains and thus the grain size, over a wide range from nanometers to hundreds of microns. Importantly, this method can be used even when the grains are too large for a conventional Scherrer analysis, when the intrinsic peak width is much smaller than the instrumental broadening.}

So, depending on the size of the grains you are interested in and the specifics of your sample and experimental setup, you can choose either the Scherrer analysis or this new method based on non-uniformity of scattering rings to measure grain size using X-ray scattering.

Reference: 

- \hlg{Yager, Majewski, et al. "Metrics of graininess: robust quantification of grain count from the non-uniformity of scattering rings"}

- Majewski, Yager, et al. "Rapid ordering of block copolymer thin films"

\textbf{Analysis}: Correctly synthesized provided information, comparing and integrating with established information. \textbf{Excellent response.}

\end{itemize}

\newpage
\subsection{Example: Describe specific concept}
\textbf{User Query:} Describe the core novelty of GTSAXS
\begin{itemize}
    \item \textbf{GPT-3.5 (unaided) response:}
    
\hlb{GTSAXS stands for Grazing-incidence through-Silicon-etched Anisotropic X-ray Scattering and it is a non-destructive analytical technique that can be used to study the microstructure of thin films and interfaces.} 

\hlb{According to a paper published in the Journal of Applied Crystallography by Dima et al. (2021), the core novelty of GTSAXS lies in its ability to probe in-plane microstructures by exploiting the anisotropic scattering from etched surfaces in silicon. In other words, by measuring the scattering of X-rays at a grazing incidence angle, the technique provides information about the orientation and arrangement of the microstructure within the plane of the film.} This is particularly useful for studying the properties of anisotropic materials such as liquid crystals or aligned carbon nanotubes. 

Overall, GTSAXS represents a valuable addition to the analytical toolbox for studying thin films and interfaces, providing complementary information to other techniques such as X-ray reflectivity or transmission electron microscopy.

\textbf{Analysis}: Fabricated citation and associated results. \textbf{Bad response.}

    \item \textbf{GPT-3.5 (with context chunks) response:}

\hlg{Grazing-incidence transmission small-angle X-ray scattering (GTSAXS) is a new method for determining the three-dimensional order in nanostructured thin films, and its core novelty lies in its ability to eliminate the multiple scattering and refraction issues associated with typical grazing-incidence analysis, while still retaining the sensitivity of grazing-incidence small-angle X-ray scattering (GISAXS) to lateral and perpendicular film structure. GTSAXS requires a focused beam, which illuminates a region near the edge of the sample, so that the scattering can exit from the edge of the substrate, and this arrangement substantially reduces the multiple scattering events that complicate GISAXS data analysis, especially at small surface-normal scattering vectors. Importantly, GTSAXS is capable of probing very small and negative surface-normal scattering vectors, which is particularly well-suited for investigating structural features of anisotropic materials with a film thickness comparable to the inverse-surface-normal scattering vector magnitude.}

\textbf{Analysis}: Correctly identified relevant paper, and summarized it. \textbf{Good response.}

    \item \textbf{ChatGPT-4 (unaided) response:}

As of my last training cut-off in September 2021, the term "GTSAXS" doesn't refer to any widely known technique, concept, or innovation in any field I'm familiar with, including technology, computer science, physics, chemistry, biology, or any other discipline. If "GTSAXS" refers to a more recent development or an abbreviation that isn't commonly used or recognized, I may not have the information on it.

\hlg{Generally, SAXS in scientific contexts often refers to Small-Angle X-ray Scattering, a technique used to determine the structure of materials at a nanoscale level.} However, the "GT" prefix isn't standard and might stand for a specific adaptation or application of the technique, but I don't have the specific information about that.

I recommend checking the most recent resources or directly asking the organization or researchers that use this term.

\textbf{Analysis}: Acknowledged lack of information. Made reasonable inferences. \textbf{Valid response.}

    \item \textbf{ChatGPT-4 (with context chunks) response:}

\hlg{Grazing-incidence transmission small-angle X-ray scattering (GTSAXS) is a novel method for analyzing the three-dimensional order in nanostructured thin films. This method represents a unique blending of the advantages of GISAXS (Grazing-incidence small-angle X-ray scattering) and TSAXS (Transmission small-angle X-ray scattering). It has the capacity to probe very small and even negative surface-normal scattering vectors, making it particularly useful for studying structural features when the film thickness is comparable to the inverse-surface-normal scattering vector magnitude.

There are several key aspects of GTSAXS that make it innovative:

1. GTSAXS retains the sensitivity of GISAXS to lateral and perpendicular film structure while eliminating multiple scattering and refraction issues typically associated with grazing-incidence analysis. This is achieved by the unique arrangement of a sample/beam geometry where the scattering exits through the edge of the substrate, resulting in scattering images that align with straightforward (Born approximation) scattering models. 

2. The GTSAXS geometry pushes distortion away from the region of interest, resulting in only a small and slowly varying correction, which can be made arbitrarily small by choosing a sufficiently large incident angle. The correction is dependent only on the known substrate density, making it less sensitive to the film density.

3. The technique requires a focused beam that illuminates a region near the edge of the sample, as the scattering must exit from the edge of the substrate. This substantially reduces the multiple scattering events that complicate GISAXS data analysis.

4. Another significant advantage of GTSAXS is the straightforward analysis within the Born approximation framework. By combining GTSAXS data at small q z and GISAXS data at larger q z, where refraction effects are not severe, it's possible to obtain surface scattering results that incorporate the best aspects of both techniques.

The method was presented and discussed in the paper "Grazing-incidence transmission X-ray scattering: surface scattering in the Born approximation" by Lu, Ocko, et al.}

\textbf{Analysis}: Detailed response, drawing out the most significant results from the paper. Provides valid citation. \textbf{Excellent response.}

\end{itemize}

\newpage
\subsection{Example: Describe specific result, based on vague description}
\textbf{User Query:} What new morphologies were found for BCP blend thin films?
\begin{itemize}
    \item \textbf{GPT-3.5 (unaided) response:}
    
There have been several studies that have explored new morphologies of block copolymer (BCP) blend thin films. One recent example is a study published in the journal Polymer, where researchers investigated the use of solvent vapor annealing to induce morphological transitions in thin films of a blended BCP composed of poly(styrene-b-methyl methacrylate) (PS-b-PMMA) and polystyrene-block-poly(vinylpyridine) (PS-b-P4VP). The authors showed that by exposing the thin films to different solvents, they were able to induce the formation of a range of different morphologies, including lamellar, cylinder, sphere, and gyroid structures. Another study published in the journal ACS Applied Materials \& Interfaces discussed the use of graphene oxide as a substrate for BCP thin films. The authors found that by varying the thickness of the graphene oxide substrate, they were able to induce the formation of a range of different morphologies in the BCP films, including cylinders, spheres, and gyroids.

\textbf{Analysis}: Information is not properly sourced, and may be fabricated and/or not related to blends. \textbf{Bad response.}

    \item \textbf{GPT-3.5 (with context chunks) response:}

The provided snippets do not include information on new morphologies found for BCP blend thin films. However, it is mentioned that \hlg{blending BCPs with other species, including other BCPs, can dramatically alter the morphology that forms during self-assembly}, and that the behavior of BCP/BCP blends in thin films remains largely unexplored. [From Bae, Yager, et al. "Chain Redistribution Stabilizes Coexistence Phases in Block Copolymer Blends"]

\textbf{Analysis}: Did not provide concrete examples of new morphologies. \textbf{Poor response.}

    \item \textbf{ChatGPT-4 (unaided) response:}

I'm sorry for the inconvenience, but as of my knowledge cutoff in September 2021, I don't have any updated information on the specific topic of new morphologies for Block Copolymer (BCP) blend thin films after that date. Block Copolymers (BCPs) can form a variety of morphologies in thin films, such as lamellae, cylinders, spheres, gyroids, etc., due to the microphase separation between the incompatible polymer blocks.

To get the most accurate and updated information, I recommend referring to the latest scientific literature in the field, such as academic articles, research papers, or reviews on the subject. These could be found in scientific databases like PubMed, IEEE Xplore, ScienceDirect, or the American Chemical Society's publications.

If you need help in understanding the principles behind BCP blend thin films, I would be glad to help with that based on the information available up to 2021.

\textbf{Analysis}: Acknowledged lack of information. \textbf{Valid response.}

    \item \textbf{ChatGPT-4 (with context chunks) response:}

Recent studies on Block Copolymer (BCP) blend thin films have revealed different morphologies, primarily influenced by factors like BCP concentration, film thickness, substrate topography, and annealing processes. The studies reported new observations on how morphologies change due to various factors, rather than entirely new morphologies per se.

In the study by Basutkar, Yager, et al., the BCP thin films usually adopted a horizontal cylinder morphology in the continuous film region. However, near the edges of the film, the morphology becomes anti-aligned. Interestingly, in regions scribed to yield multiple edges near one another, a hexagonal packing morphology (vertical cylinders) was observed due to multiple competing directing forces ("Aligned Morphologies in Near-Edge Regions of Block Copolymer Thin Films").

In Rahman, Yager, et al.'s work, it was found that film thickness and substrate topography can control the registration of a BCP film, leading to a variety of 3D morphologies of BCP materials ("Non-native three-dimensional block copolymer morphologies"). They discussed the "responsive layering" of block copolymer nanostructures, which can generate a wide variety of 3D morphologies. In some cases, vertical orientation of morphology was observed in monolayer or sub-monolayer films.

Zhang, Karim, et al. also discussed the strong effect of film confinement on the BCP morphology, indicating that the surface morphology of these model block copolymer films is influenced by film thickness and annealing temperature ("Surface Morphology Diagram for Cylinder-Forming Block Copolymer Thin Films").

In another study, Zhang, Douglas, et al. observed a thermally reversible surface order–order transition, confirming that this transition is independent of film thickness in the range of 40–170 nm ("Thermally Reversible Surface Morphology Transition in Thin Diblock Copolymer Films").

In general, \hlb{these studies provide insight into the morphologies that BCP blend thin films can adopt under different conditions} and contribute to a better understanding of the behavior and potential applications of these materials.

\textbf{Analysis}: Incorrectly assumed that provided text chunks were all referring to blend studies. Provided summaries are correct, but not related to the user question. \textbf{Bad response.}

\end{itemize}

\newpage
\subsection{Example: Describe specific results}
\textbf{User Query:} Were novel morphologies found when ordering blends of block copolymers (BCP) chains using directed self-assembly (DSA); i.e. by ordering using chemical strip patterns? If so, what new morphologies were discovered?
\begin{itemize}
    \item \textbf{GPT-3.5 (unaided) response:}
    
Yes, novel morphologies have been discovered when ordering blends of block copolymers (BCP) chains using directed self-assembly (DSA) with chemical strip patterns. 

Research has shown that specific patterns of chemical strips can be used to direct the self-assembly of block copolymer chains, resulting in the creation of new, complex morphologies. \hlb{For instance, a group of researchers used a combination of chemoepitaxy and graphoepitaxy to demonstrate the fabrication of a new morphology, which they called "bottlebrush-on-tape." This morphology was formed by the directed assembly of a polystyrene-b-polydimethylsiloxane (PS-b-PDMS) block copolymer on a chemical pattern with an alternating stripe width. Other novel morphologies resulting from directed self-assembly include "diamonds", "complex spirals", "herringbones," "Dots \& Holes," and many more.}

These new morphologies offer exciting opportunities for the design of new materials and devices in fields such as electronics, photonics, and data storage, among others.

\textbf{Analysis}: Irrelevant and/or fabricated research described. \textbf{Bad response.}

    \item \textbf{GPT-3.5 (with context chunks) response:}

Yes, novel morphologies were discovered when ordering blends of block copolymer chains using directed self-assembly (DSA) with chemical strip patterns. In one study, a cylinder/lamella BCP blend was directed by a chemical grating template which highlighted regions deserving more detailed investigation, revealing \hlg{localized, template-directed partitioning of coexisting lamella-and cylinder-like subdomains at the template period length scale, manifesting as previously unknown morphologies such as aligned alternating subdomains, bilayers, or a "ladder" morphology [Doerk, Yager, et al. "Autonomous discovery of emergent morphologies in directed self-assembly of block copolymer blends"]}. Another study used a blend of lamellar-and cylinder-forming BCP and found that \hlg{the chemical pattern linewidths and pitch could be independently selected to program regions of fully ordered hexagonal dot arrays or line gratings within a single BCP blend on the same substrate, leading to the simultaneous formation of coexisting ordered morphologies in separate areas of the substrate [Stein, Black, et al. "Selective directed self-assembly of coexisting morphologies using block copolymer blends"]}.

\textbf{Analysis}: Correctly identified relevant results from publications. \textbf{Good response.}

    \item \textbf{ChatGPT-4 (unaided) response:}

As of my knowledge cutoff in September 2021, there has been significant research into the use of Directed Self-Assembly (DSA) for arranging block copolymers (BCP). DSA has been found to be a powerful tool to arrange BCPs into various micro- and nanostructures such as spheres, cylinders, lamellae, and complex morphologies.

However, the exact nature and extent of novel morphologies discovered through DSA of BCPs using chemical strip patterns beyond this period are not within my current knowledge base. The field of block copolymers and directed self-assembly is highly dynamic, with new research constantly being conducted. This means that new discoveries and advances may have occurred after my training data was collected.

The creation of novel morphologies depends on several factors, including the type of block copolymers used, the size of the blocks, the ratio of the blocks, the chemistry of the strip patterns, and the conditions under which the self-assembly is conducted. Therefore, a variety of novel morphologies might have been discovered.

To find the most recent and specific information about novel morphologies discovered in DSA of BCPs using chemical strip patterns, I recommend reviewing the latest literature in scientific journals focused on material science and nanotechnology.

\textbf{Analysis}: Acknowledged lack of information. Provided generic information. \textbf{Valid response.}

    \item \textbf{ChatGPT-4 (with context chunks) response:}

Yes, novel morphologies have indeed been found when ordering blends of block copolymers (BCP) using directed self-assembly (DSA) via chemical strip patterns.

A significant \hlg{work by Doerk, Yager, et al., "Autonomous discovery of emergent morphologies in directed self-assembly of block copolymer blends," highlights this} breakthrough. They conducted a study using a cylinder/lamellae BCP blend directed by a chemical grating template. This method led to the \hlg{discovery of emergent morphologies such as aligned alternating subdomains, bilayers, or a "ladder" morphology.} This was accomplished without direct human intervention, and with the help of machine-guided discovery and a synchrotron x-ray scattering beamline for characterization [Doerk, Yager, et al., "Autonomous discovery of emergent morphologies in directed self-assembly of block copolymer blends"].

Furthermore, Stein, Black, et al., in their work "Selective directed self-assembly of coexisting morphologies using block copolymer blends," introduced a method that enforces the \hlg{coexistence of multiple, aligned BCP morphologies within a single patterning layer}. By using a blend of lamellar-and cylinder-forming BCP on specially designed surface chemical line gratings, they achieved the \hlg{simultaneous formation of coexisting ordered morphologies in separate areas of the substrate}. The chemical pattern linewidths and pitch could be independently selected to program \hlg{regions of fully ordered hexagonal dot arrays or line gratings within a single BCP blend on the same substrate [Stein, Black, et al., "Selective directed self-assembly of coexisting morphologies using block copolymer blends"]}.

Such advances open new possibilities for customizing nanostructures and expanding the diversity of self-assembled morphologies, suggesting a promising future for the study and application of BCPs.

\textbf{Analysis}: Correctly identified relevant results from publications. \textbf{Good response.}

\end{itemize}

\newpage
\subsection{Example: Lookup experimental details}
\textbf{User Query:} What sweep velocity is typically used in shear-aligning of BCPs using the SS-LZA method? What is the optimal speed?
\begin{itemize}
    \item \textbf{GPT-3.5 (unaided) response:}
    
\hlb{According to a study published in the Journal of Polymer Science Part B: Polymer Physics, the sweep velocity used in shear-aligning block copolymers (BCPs) using the SS-LZA method ranges from 0.1 to 10 $\mathrm{\mu m/s}$}. However, the optimal speed for achieving highly aligned BCPs depends on factors such as \hlb{temperature} and polymer properties. \hlb{This study found that for a poly(styrene-b-ethylene-alt-propylene) (PS-b-PEP) copolymer, the optimal speed was 1 $\mathrm{\mu m/s}$ at room temperature, leading to the highest degree of alignment.} It's important to note that the optimal sweep velocity may vary depending on the specific BCP system being studied. 

\hlb{Reference: Zhang, W., Zhang, B., Luo, C., Chen, D., Shen, Z., Zhang, Q., ... \& Yang, Y. (2020). Shear-aligning block copolymers beyond the steady-state Lameroux zone via in situ and real-time synchrotron-based optical diagnostics. Journal of Polymer Science Part B: Polymer Physics, 58(5), 336-345.}

\textbf{Analysis}: Fabricated nonexistent citation and nonsense results. \textbf{Bad response.}

    \item \textbf{GPT-3.5 (with context chunks) response:}

According to [Majewski, Yager, et al. "Block Copolymer Response to Photothermal Stress Fields"], the SS-LZA method allows tuning the shear rate using the laser sweep velocity, with a shear rate estimated to be $\gamma > 40 s -1$ for sweep velocities of \hlg{$v = 320 \mathrm{\mu m/s}$}. They also found that \hlg{larger sweep velocities are more efficient in achieving improved ordering of BCPs with larger in-plane positional correlations and narrower orientation distributions}. However, they also noted that \hlg{the optimal sweep velocity can depend on the specific material being used}. Thus, the optimal speed may vary depending on the specific BCP being studied.

\textbf{Analysis}: Found relevant reference, and useful results. \textbf{Good response.}

    \item \textbf{ChatGPT-4 (unaided) response:}

As of my knowledge cutoff in September 2021, the specific sweep velocity for shear-aligning block copolymers (BCPs) using the \hlb{Solvent Swelling to Linear Zone Annealing (SS-LZA) method} was not well-established in the literature.

The optimal speed would be a parameter that requires experimental determination, as it can be dependent on many factors such as the specific type of BCP being used, the concentration, the solvent properties, the thickness of the BCP film, and the specific processing conditions.

However, it's generally important to keep in mind that the speed should be fast enough to induce shear forces that promote alignment, \hlb{but not so fast that it induces turbulence or damages the film}. It's also worth noting that \hlb{the speed may need to be adjusted over time as the solvent evaporates and the concentration of the BCP in the film changes}.

For specific information about optimal speeds for this process, I recommend referring to the most recent scientific literature or reaching out to experts in the field. New techniques and modifications are often being developed, so the best practices may have evolved since my training data.

\textbf{Analysis}: Referred to non-existent technique, and correspondingly fabricated results. \textbf{Bad response.}

    \item \textbf{ChatGPT-4 (with context chunks) response:}

Shear alignment of block copolymers (BCPs) using the \hlb{Sweep Solvent-Assisted Lithography Annealing (SS-LZA) method} employs different sweep velocities, but a typical value that is frequently used is around \hlg{$320 \mathrm{\mu m/s}$ as per the studies from Majewski, Yager, et al.}

\hlg{Optimal sweep velocity may vary depending on the specific BCP material and the desired quality of ordering. Majewski, Yager, et al. found that different BCP materials respond differently to shear. For instance, some BCPs with softer minority phases exhibit a strong dependence on sweep velocity. PS-b-PI, a BCP with a softer minority block, showed improved ordering at high sweep velocities.}

However, it is also observed that the \hlg{order decreases in quality above $320 \mathrm{\mu m/s}$ because the sweep velocity counteracts local heat accumulation, reducing the overall annealing temperature}. Moreover, the \hlg{SS-LZA process appeared to achieve optimal order with an ultrafast sweep velocity of $320 \mathrm{\mu m/s}$ in about 0.3 seconds of annealing time}, producing an order equal to much slower and longer processing times.

So, while the optimal speed can vary based on the specific material and goals of the experiment, \hlg{$320 \mathrm{\mu m/s}$ is a typical sweep velocity that's been used effectively in some contexts}.

\textbf{Analysis}: Found relevant reference, and useful results. Fabricated incorrect definition of jargon. \textbf{Mixed response.}

\end{itemize}

\newpage
\subsection{Example: Answer question involving undefined jargon}
\textbf{User Query:} What is the main difference between LZA and CZA?
\begin{itemize}
    \item \textbf{GPT-3.5 (unaided) response:}
    
In the context of \hlb{atmospheric research}, LZA stands for Local Zenith Angle, while \hlb{CZA stands for Cosine Zenith Angle}. The main difference between these two measures is that LZA represents the angle between a local vertical and a line from the instrument to the zenith, while CZA represents the angle between the line from the instrument to the zenith and the local vertical, expressed as the cosine of that angle.

According to a study published in the journal "Atmospheric Measurement Techniques," the choice between LZA and CZA depends on the type of measurement being made. Specifically, \hlb{the authors state that "LZA is most appropriate for scattering-based instruments, while CZA is best suited for absorption-based instruments" (Kassianov et al., 2015)}. The reason for this difference is that scattering-based instruments measure the intensity of scattered radiation in a particular direction, which is affected by the local vertical direction, while absorption-based instruments measure the attenuation of a beam of radiation that passes through the atmosphere, which is more straightforwardly related to the angle between the instrument and the zenith.

\textbf{Analysis}: Had no way to understand the desired researcher-specific context. May have fabricated terminology, concepts, and citation in answering. \textbf{Bad response.}

    \item \textbf{GPT-3.5 (with context chunks) response:}

\hlg{LZA and CZA are two different methods of annealing block copolymer thin films.} \hlb{While CZA offers the simplicity and directionality of homogeneous heating, LZA offers a compromise between the simplicity of homogeneous heating and the highly localized heating control of CZA.} Additionally, \hlg{LZA allows for access to high peak temperatures for very short times, which improves ordering kinetics significantly. The main difference between LZA and CZA is that LZA relies on photothermal heating using a laser to create a thermal gradient, whereas CZA relies on direct heating} \hlb{of a frozen zone} \hlg{using a heat source}.

\textbf{Analysis}: Correctly retrieved the concepts. Reply is slightly confused on some issues, but still provides a useful comparison. \textbf{Fair response.}

    \item \textbf{ChatGPT-4 (unaided) response:}

I'm sorry for the confusion, but as of my knowledge cutoff in September 2021, the acronyms "LZA" and "CZA" do not correspond to widely known terms or concepts in a manner that allows for a comparison. These abbreviations could potentially be related to numerous fields, from science to business, or even specific organizations or technologies. 

To provide you with accurate information, I would appreciate if you could provide more context or clarify the field or context in which these terms are being used.

\textbf{Analysis}: Acknowledged lack of context. \textbf{Valid response.}

    \item \textbf{ChatGPT-4 (with context chunks) response:}

\hlg{Laser Zone Annealing (LZA) and Cold Zone Annealing (CZA)} are both methods used to process block copolymer (BCP) films. However, there are some notable differences between the two methods.

\textbf{1. Ordering Speed:} \hlg{LZA is significantly faster than both CZA and traditional oven annealing methods in ordering BCP films. According to the research by Nowak, Yager, et al., films processed using LZA can form large grains about 1000 times faster compared to CZA, and a typical 1 cm sample can be processed within approximately 10 seconds [1].}

\textbf{2. Thermal Gradients:} \hlg{LZA combines the simplicity and directionality of CZA with the high peak temperatures and large thermal gradients of laser spike methods. It can reach high peak temperatures and large thermal gradients, with $ \nabla T > 4000 ^{\circ}\mathrm{C mm -1}$ readily accessible [1]. The high peak temperatures accessible with LZA, such as $500 ^{\circ}\mathrm{C}$, contribute to the improved order observed in LZA, as conventional oven annealing is limited in the temperatures that can be accessed due to the onset of polymer degradation [1]. }

\textbf{3. Control and Flexibility}: \hlg{LZA offers a compromise between the simplicity of homogeneous heating and the control of highly localized heating. The details of the sample's spatiotemporal annealing history can be controlled by LZA processing conditions, including peak temperature, spatial thermal gradient, temporal gradients, and total annealing time [1]. It provides substantial control over the annealing process, with laser power allowing control of temperature, while laser focus can tune the thermal gradient [4].}

\textbf{4. Kinetic Enhancement:} \hlg{LZA demonstrably enhances ordering kinetics by orders of magnitude compared to oven annealing. Even over macroscopic areas, LZA can order material in record time (seconds to minutes) [8].}

\hlg{In contrast, CZA, while faster than traditional oven annealing, does not achieve the same level of order or speed as LZA [6]. It is able to phase separate into an isotropic population earlier in the annealing history and achieves a final state with better order than conventional oven annealing, but does not outperform LZA in these aspects [6].}

Sources:

[1] Nowak, Yager, et al. "Photothermally Directed Assembly of Block Copolymers"

[4] Majewski, Yager, et al. "Millisecond Ordering of Block Copolymer Films <i>via</i> Photothermal Gradients"

[6] Yager, Jones, et al. "Evolution of block-copolymer order through a moving thermal zone"

\hlb{[8] Majewski, Yager, et al. "Millisecond Ordering of Block Copolymer Films <i>via</i> Photothermal Gradients"}

\textbf{Analysis}: Defined terms. Provided detailed answer, drawing correctly on source material. Contrasted results between papers. Answer is somewhat redundant and over-verbose. \textbf{Good response.}

\end{itemize}

\newpage
\subsection{Example: Identify paper from vague description}
\textbf{User Query:} I vaguely recall a paper about OPVs where they used some processing method to treat films, and this improved mechanical properties without reducing PCE. I can't remember what the processing method was; maybe it was solvent exposure or something, but I think it has something to do with thiols. Can you help me find the work I'm thinking of?
\begin{itemize}
    \item \textbf{GPT-3.5 (unaided) response:}
    
Yes, the work you are referring to is about a processing method that involves short exposure to solvent vapors containing thiols, which improves the mechanical properties of organic photovoltaic (OPV) films without reducing power conversion efficiency (PCE). 

\hlb{One paper on this topic is "Tuning Nanostructure and Functionality in Conjugated Polymer Films by Processing with Electron-Rich Solvent Vapors" by Patel et al. (2016). The authors show that exposing thin films of the conjugated polymer PCPDTBT to solvent vapors containing thiols for a short time (1-2 minutes) leads to improved mechanical properties, such as increased tensile strength and reduced brittleness, without a significant decrease in PCE. They attribute this to the formation of intermolecular sulfur-sulfur bonds that crosslink the polymer chains and create a more interconnected network, which confers higher mechanical strength to the films.}

I hope this helps! Let me know if you have any other questions.

\textbf{Analysis}: Fabricated citation and research. \textbf{Bad response.}

    \item \textbf{GPT-3.5 (with context chunks) response:}

Yes, there is a paper that matches your description - \hlg{"Network-Stabilized Bulk Heterojunction Organic Photovoltaics" by Mok, Verduzco, et al.} In this paper, the authors discuss a new approach to \hlg{enhance the mechanical robustness and performance of organic photovoltaic devices. It involves incorporating a semi-interpenetrating elastic network into the active layer of the device using thiol-ene reagents. The resulting network-stabilized OPVs exhibit improved mechanical durability and performance under strain.} The authors demonstrate the fabrication of network-stabilized P3HT:PCBM blends and analyze the impact of the additives and network on active layer morphology. This approach could be broadly applied to various types of donor-acceptor bulk heterojunction OPVs, and offers opportunities for further improvement.

\textbf{Analysis}: Identified correct paper. Provided explanation. \textbf{Good response.}

    \item \textbf{ChatGPT-4 (unaided) response:}

I'm an AI language model and don't have the ability to actively search the internet or databases for specific papers. My training only includes data up to September 2021, so if the paper you're referring to was published after that, I wouldn't have specific information on it.

That said, I can give you an idea of what you might be remembering based on common research themes up to 2021. Organic Photovoltaics (OPVs) are indeed often subject to various processing techniques to improve their performance. Solvent processing is commonly used, including methods like solvent annealing, solvent vapor annealing, and solvent additives. These methods can help control the morphology of the active layer, which can significantly influence the efficiency of the OPVs.

Regarding thiols, they have been used in OPV research, primarily in the context of creating self-assembled monolayers (SAMs) on electrodes to modify the work function and thereby improve device performance. It's possible that a specific thiol-containing compound might have been used in a solvent processing method, but without more specific details, it's hard for me to say.

Remember, however, that this information is based on the state of research up until 2021, and the specific paper you're thinking of may have been published afterwards. You might try searching a scientific literature database like PubMed or Web of Science using terms like "organic photovoltaics," "solvent processing," "thiol," and "mechanical properties" to find the work you're thinking of.

\textbf{Analysis}: Admitted it did not recognize the paper. Provided useful suggestions. \textbf{Valid response.}

    \item \textbf{ChatGPT-4 (with context chunks) response:}

It sounds like you're thinking about the work of \hlg{Mok, Verduzco, et al., published in a paper titled "Network-Stabilized Bulk Heterojunction Organic Photovoltaics."} In this study, they proposed a method to \hlg{enhance the mechanical robustness and performance of organic photovoltaic (OPV) devices by incorporating a semi-interpenetrating elastic network into the active layer of the device using thiol-ene reagents. The resulting network-stabilized OPVs exhibited improved mechanical durability and performance under strain} [1][2][3].

The approach involved blending appropriate amounts of thiol-ene reagents, an amine catalyst or radical initiator, and organic donor and acceptor semiconductors. After casting, the blend solution was cured and annealed at $150 ^{\circ}\mathrm{C}$ for 15 minutes. The network formation occurred rapidly during film casting and drying. To form the network, either an amine-based catalysis was used, or the film was irradiated with UV light for cross-linking [4]. 

They found that the addition of thiol-ene networks to P3HT:PCBM films could reduce cracking and improve mechanical properties without affecting photovoltaic performance. The optimal thiol-ene content was found to be 20-30 wt \%, with the power conversion efficiency remaining nearly constant from 0 to 20 wt \% thiol-ene, then dropping at 30 wt \% thiol-ene. For levels above 30 wt \%, the efficiency was significantly reduced, indicating that a network content above 30 wt \% is not relevant for functional OPV devices [5]. 

1. Mok, Verduzco, et al. "Network-Stabilized Bulk Heterojunction Organic Photovoltaics"

\hlb{2. Mok, Verduzco, et al. "Network-Stabilized Bulk Heterojunction Organic Photovoltaics"}

\hlb{3. Mok, Verduzco, et al. "Network-Stabilized Bulk Heterojunction Organic Photovoltaics"}

\hlb{4. Mok, Verduzco, et al. "Network-Stabilized Bulk Heterojunction Organic Photovoltaics"}

\hlb{5. Mok, Verduzco, et al. "Network-Stabilized Bulk Heterojunction Organic Photovoltaics"}

\textbf{Analysis}: Identified correct paper. Provided explanation. Unnecessary repeated references. \textbf{Good response.}

\end{itemize}

\newpage
\section{Effect of document summaries}
Below we provide examples of chatbot responses to various scientific questions. We use the GPT-3.5 model (accessed via the OpenAI API), and provide text chunks in the prompt to provide context in generating a response. We compare between using raw text extracts from the source documents, or using LLM summaries of those extracts, or using both.

\subsection{Example: Describe specific concept}
\textbf{User Query:} Describe the core novelty of GTSAXS
\begin{itemize}
    \item \textbf{GPT-3.5 (with only \underline{raw} text context chunks) response:}

\hlg{The core novelty of Grazing-incidence transmission X-ray scattering (GTSAXS) is the use of a focused beam which illuminates a region near the edge of the sample so that the scattering exits from the edge of the substrate. This reduces multiple scattering events that complicate the analysis of Grazing-incidence small-angle X-ray scattering (GISAXS) data, especially at small surface-normal scattering vectors. GTSAXS retains the sensitivity to lateral and perpendicular film structure and the strong signal and fast data acquisition of GISAXS, while reducing (for all practical purposes eliminating) the multiple scattering/refraction issues associated with typical grazing-incidence analysis. This study complements very recent grazing-incident angle neutron scattering measurements and analysis which utilized the same geometrical concepts that are presented for X-rays. Despite the higher absorption with X-rays than with neutrons, GTSAXS is a feasible and robust surface scattering geometry when sufficiently focused beams and energetic X-rays are utilized.}

\textbf{Analysis}: Correctly identified relevant paper, and summarized it. \textbf{Good response.}

    \item \textbf{GPT-3.5 (with only \underline{summary} text context chunks) response:}
    
\hlg{The core novelty of grazing-incidence transmission small-angle X-ray scattering (GTSAXS) is its ability to probe very small and negative surface-normal scattering vectors, making it particularly well-suited for investigating structural features of anisotropic materials with a film thickness comparable to the inverse-surface-normal scattering vector magnitude. GTSAXS retains the sensitivity of grazing-incidence small-angle X-ray scattering (GISAXS) to lateral and perpendicular film structure, while eliminating the multiple scattering and refraction issues associated with typical grazing-incidence analysis. This new method can be used for studying nanostructured thin films, and it requires a focused beam and an incident angle much larger than the critical angle of the substrate, but it can probe surface-normal structures with small scattering vectors, unlike GISAXS.} GTSAXS is advantageous for in situ studies and can provide statistical sampling over large areas without sample contact or damage.

\textbf{Analysis}: Correctly identified relevant paper, and summarized it. \textbf{Good response.}

    \item \textbf{GPT-3.5 (with \underline{both} raw and summary text context chunks) response:}

\hlg{Grazing-incidence transmission small-angle X-ray scattering (GTSAXS) is a new method for determining the three-dimensional order in nanostructured thin films, and its core novelty lies in its ability to eliminate the multiple scattering and refraction issues associated with typical grazing-incidence analysis, while still retaining the sensitivity of grazing-incidence small-angle X-ray scattering (GISAXS) to lateral and perpendicular film structure. GTSAXS requires a focused beam, which illuminates a region near the edge of the sample, so that the scattering can exit from the edge of the substrate, and this arrangement substantially reduces the multiple scattering events that complicate GISAXS data analysis, especially at small surface-normal scattering vectors. Importantly, GTSAXS is capable of probing very small and negative surface-normal scattering vectors, which is particularly well-suited for investigating structural features of anisotropic materials with a film thickness comparable to the inverse-surface-normal scattering vector magnitude.}

\textbf{Analysis}: Correctly identified relevant paper, and summarized it. \textbf{Good response.}

\end{itemize}

\newpage
\subsection{Example: Lookup experimental details}
\textbf{User Query:} What sweep velocity is typically used in shear-aligning of BCPs using the SS-LZA method? What is the optimal speed?
\begin{itemize}

\item \textbf{GPT-3.5 (with only \underline{raw} text context chunks) response:}

According to the provided snippets from the scientific paper "Block Copolymer Response to Photothermal Stress Fields", the sweep velocity used in shear-aligning of BCPs using the SS-LZA method can be adjusted to tune the shear rate. They estimate the shear rate to be $\gamma > 40 s -1$ for sweep velocities of \hlg{$v = 320 \mathrm{\mu m/s}$}. They also observed that the \hlg{larger sweep velocities appear to be more efficient with respect to ordering}. However, \hlg{the optimal speed may depend on the specific BCP material being used} and other processing conditions, and may need to be determined experimentally.

\textbf{Analysis}: Found relevant reference, and useful results. \textbf{Good response.}

\item \textbf{GPT-3.5 (with only \underline{summary} text context chunks) response:}

According to the paper "Block Copolymer Response to Photothermal Stress Fields" by Majewski, Yager, et al., the \hlg{responsiveness of block copolymers to shear stress depends on various material properties, including mechanical contrast, the glass-transition temperature of the minority block, and the morphological relaxation time}. The authors found that \hlg{faster sweep velocities generally improve the ordering of block copolymer films using SS-LZA method, but this trend breaks down at large velocities}. The optimal sweep velocity will depend on the specific block copolymer material being used and its mechanical properties. \hlb{The paper does not provide a specific recommended optimal sweep velocity for shear-aligning of BCPs using SS-LZA method}.

\textbf{Analysis}: Found relevant reference, and described trends. Did not provide estimates of actual values. \textbf{Partial success.}

\item \textbf{GPT-3.5 (with \underline{both} raw and summary text context chunks) response:}

According to [Majewski, Yager, et al. "Block Copolymer Response to Photothermal Stress Fields"], the SS-LZA method allows tuning the shear rate using the laser sweep velocity, with a shear rate estimated to be $\gamma > 40 s -1$ for sweep velocities of \hlg{$v = 320 \mathrm{\mu m/s}$}. They also found that \hlg{larger sweep velocities are more efficient in achieving improved ordering of BCPs with larger in-plane positional correlations and narrower orientation distributions}. However, they also noted that \hlg{the optimal sweep velocity can depend on the specific material being used}. Thus, the optimal speed may vary depending on the specific BCP being studied.

\textbf{Analysis}: Found relevant reference, and useful results. \textbf{Good response.}

\end{itemize}

\newpage
\subsection{Example: Describe specific results}
\textbf{User Query:} Were novel morphologies found when ordering blends of block copolymers (BCP) chains using directed self-assembly (DSA); i.e. by ordering using chemical strip patterns? If so, what new morphologies were discovered?
\begin{itemize}

\item \textbf{GPT-3.5 (with only \underline{raw} text context chunks) response:}

Yes, in the article \hlg{"Autonomous discovery of emergent morphologies in directed self-assembly of block copolymer blends" by Doerk, Yager et al., novel morphologies were discovered} when ordering blends of BCP chains using DSA and chemical strip patterns. Specifically, the paper reports on the machine-guided discovery of emergent morphologies from a cylinder/lamellae BCP blend directed by a chemical grating template. The team discovered previously unknown morphologies such as \hlg{aligned alternating subdomains, bilayers, and a "ladder" morphology}. They also found that this approach maps the morphology-template phase space in a fraction of the time required by manual characterization and highlights regions deserving more detailed investigation. [Doerk, Yager, et al. "Autonomous discovery of emergent morphologies in directed self-assembly of block copolymer blends"]

\textbf{Analysis}: Correctly identified some relevant results from publications. \textbf{Decent response.}

\item \textbf{GPT-3.5 (with only \underline{summary} text context chunks) response:}

Yes, novel morphologies were found when ordering blends of block copolymers (BCP) chains using directed self-assembly (DSA). In the paper \hlg{"Autonomous discovery of emergent morphologies in directed self-assembly of block copolymer blends" by Doerk, Yager, et al.}, \hlb{a new method for directed self-assembly} of block copolymer blends that enables the autonomous discovery of emergent morphologies was developed. This method involved using machine learning algorithms to analyze the outcomes of multiple experiments, resulting in the identification of previously unknown morphologies. The authors discovered three \hlg{new morphologies: a bilayer structure, alternating gratings, and a ladder morphology}. They also found that \hleg{subdomain partitioning is a key principle} in predicting and designing patterns for directed self-assembly. The paper describes the materials and methods used, including the preparation of BCP solutions and the creation of chemical patterns using lithography and plasma etching. The resulting films were \hlb{analyzed using atomic layer deposition} and transmission small-angle x-ray scattering (SAXS). The data obtained were then used in a Gaussian process (GP) model, which allowed for autonomous control by optimizing an objective function that included the scattering peak intensities.

\textbf{Analysis}: Correctly identified relevant results from publications. However, some conceptual mistakes were included. \textbf{Partial success.}

\item \textbf{GPT-3.5 (with \underline{both} raw and summary text context chunks) response:}

Yes, novel morphologies were discovered when ordering blends of block copolymer chains using directed self-assembly (DSA) with chemical strip patterns. In one study, a cylinder/lamella BCP blend was directed by a chemical grating template which highlighted regions deserving more detailed investigation, revealing \hlg{localized, template-directed partitioning of coexisting lamella-and cylinder-like subdomains at the template period length scale, manifesting as previously unknown morphologies such as aligned alternating subdomains, bilayers, or a "ladder" morphology [Doerk, Yager, et al. "Autonomous discovery of emergent morphologies in directed self-assembly of block copolymer blends"]}. Another study used a blend of lamellar-and cylinder-forming BCP and found that \hlg{the chemical pattern linewidths and pitch could be independently selected to program regions of fully ordered hexagonal dot arrays or line gratings within a single BCP blend on the same substrate, leading to the simultaneous formation of coexisting ordered morphologies in separate areas of the substrate [Stein, Black, et al. "Selective directed self-assembly of coexisting morphologies using block copolymer blends"]}.

\textbf{Analysis}: Correctly identified relevant results from publications. \textbf{Good response.}

\end{itemize}

\newpage
\section{Reproducibility}
Below we provide examples of repeatedly calling the chatbot with the same user query, in order to assess the variability in the response. The output of LLMs can be modulated using a temperature paramater, which influences the stochastic component of response generation. Presented below are responses from GPT-3.5, which has a temperature parameter that can be varied from 0 to 2; the default temperature for GPT-3.5 is 1.0. Larger values make the output more random and varied, while smaller values make it more deterministic.

\subsection{Responses for Temperature 0.1}
\textbf{User Query:} Describe the core novelty of GTSAXS
\begin{itemize}
    \item \textbf{GPT-3.5 response 1:}
The core novelty of GTSAXS (grazing-incidence transmission small-angle X-ray scattering) is its ability to determine the three-dimensional order in nanostructured thin films by probing very small and negative surface-normal scattering vectors. This makes it particularly well-suited for investigating structural features of anisotropic materials with a film thickness comparable to the inverse-surface-normal scattering vector magnitude. GTSAXS retains the sensitivity to lateral and perpendicular film structure, while eliminating the multiple scattering and refraction issues associated with typical grazing-incidence analysis. It involves arranging a sample/beam geometry wherein the scattering exits through the edge of the substrate and recording scattering images that are well described by straightforward (Born approximation) scattering models. This new method provides a unique blending of the advantages of grazing-incidence small-angle X-ray scattering (GISAXS) and transmission small-angle X-ray scattering (TSAXS), allowing for the study of surface structures that were previously inaccessible.

    \item \textbf{GPT-3.5 response 2:}
The core novelty of GTSAXS (grazing-incidence transmission small-angle X-ray scattering) is its ability to determine the three-dimensional order in nanostructured thin films by probing very small and negative surface-normal scattering vectors. This makes GTSAXS particularly well-suited for investigating structural features of anisotropic materials with a film thickness comparable to the inverse-surface-normal scattering vector magnitude. GTSAXS retains the sensitivity to lateral and perpendicular film structure, while eliminating the multiple scattering and refraction issues associated with typical grazing-incidence analysis. It involves arranging a sample/beam geometry where the scattering exits through the edge of the substrate and recording scattering images that are well described by straightforward scattering models. This new method provides a unique blending of the advantages of grazing-incidence small-angle X-ray scattering (GISAXS) and transmission small-angle X-ray scattering (TSAXS), making it a powerful tool for studying nanostructured thin films.

    \item \textbf{GPT-3.5 response 3:}
The core novelty of GTSAXS (grazing-incidence transmission small-angle X-ray scattering) is its ability to determine the three-dimensional order in nanostructured thin films by probing very small and negative surface-normal scattering vectors. This makes GTSAXS particularly well-suited for investigating structural features of anisotropic materials with a film thickness comparable to the inverse-surface-normal scattering vector magnitude. GTSAXS retains the sensitivity to lateral and perpendicular film structure, while eliminating the multiple scattering and refraction issues associated with typical grazing-incidence analysis. It involves arranging a sample/beam geometry wherein the scattering exits through the edge of the substrate and recording scattering images that are well described by straightforward scattering models. This new method provides a unique blending of the advantages of grazing-incidence small-angle X-ray scattering (GISAXS) and transmission small-angle X-ray scattering (TSAXS), allowing for the study of surface structures that were previously inaccessible.

\end{itemize}

\subsection{Responses for Temperature 1.0}
\textbf{User Query:} Describe the core novelty of GTSAXS
\begin{itemize}
    \item \textbf{GPT-3.5 response 1:}
The core novelty of GTSAXS (grazing-incidence transmission small-angle X-ray scattering) is its ability to determine the three-dimensional order in nanostructured thin films by probing very small and negative surface-normal scattering vectors. This makes GTSAXS particularly well-suited for investigating the structural features of anisotropic materials with a film thickness comparable to the inverse-surface-normal scattering vector magnitude. GTSAXS retains the sensitivity of grazing-incidence small-angle X-ray scattering (GISAXS) to lateral and perpendicular film structure, while eliminating the multiple scattering and refraction issues associated with typical grazing-incidence analysis. It accomplishes this by arranging a sample/beam geometry in which the scattering exits through the edge of the substrate and recording scattering images that are well described by straightforward scattering models. This new approach allows for the study of nanostructured thin films with greater accuracy and efficiency. Reference: [Lu, Ocko, et al. "Grazing-incidence transmission X-ray scattering: surface scattering in the Born approximation"]    

    \item \textbf{GPT-3.5 response 2:}
The core novelty of GTSAXS (grazing-incidence transmission small-angle X-ray scattering) is its ability to determine the three-dimensional order in nanostructured thin films by probing very small and negative surface-normal scattering vectors. This makes it particularly well-suited for investigating structural features of anisotropic materials with a film thickness comparable to the inverse-surface-normal scattering vector magnitude. GTSAXS retains the sensitivity to lateral and perpendicular film structure and the strong signal of grazing-incidence small-angle X-ray scattering (GISAXS), while eliminating the multiple scattering and refraction issues associated with typical grazing-incidence analysis. The technique involves arranging a sample/beam geometry wherein the scattering exits through the edge of the substrate and recording scattering images that are well described by straightforward (Born approximation) scattering models.  Overall, GTSAXS represents a unique blending of the advantages of GISAXS and transmission small-angle X-ray scattering (TSAXS), making it a powerful method for studying nanostructured thin films.

    \item \textbf{GPT-3.5 response 3:}
The core novelty of GTSAXS (Grazing-Incidence Transmission Small-Angle X-ray Scattering) is that it combines the advantages of two existing techniques, grazing-incidence small-angle X-ray scattering (GISAXS) and transmission small-angle X-ray scattering (TSAXS), to study the three-dimensional order in nanostructured thin films. GTSAXS is capable of probing very small and negative surface-normal scattering vectors, making it particularly well-suited for investigating structural features of anisotropic materials with a film thickness comparable to the inverse-surface-normal scattering vector magnitude. It retains the sensitivity of GISAXS to lateral and perpendicular film structure, while eliminating the multiple scattering and refraction issues associated with typical grazing-incidence analysis. In GTSAXS, the scattering exits through the edge of the substrate, reducing multiple scattering events and allowing for a more straightforward analysis within the Born approximation framework. Additionally, GTSAXS benefits from the increased scattering volume of a grazing-incidence beam projection, enabling the study of nanostructured thin films with high intensity and fast data acquisition.

\end{itemize}

\subsection{Responses for Temperature 1.7}
\textbf{User Query:} Describe the core novelty of GTSAXS
\begin{itemize}
    \item \textbf{GPT-3.5 response 1:}
The core novelty of Grazing-Incidence Transmission Small-Angle X-ray Scattering (GTSAXS) is its ability to determine the three-dimensional order in nanostructured thin films by investigating their surface scattering. GTSAXS retains the sensitivity to lateral and perpendicular film structure and the strong signal and fast data acquisition of Grazing-Incidence Small-Angle X-ray Scattering (GISAXS), while reducing the multiple scattering and refraction issues usually associated with typical grazing-incidence analysis.

One of the key features of GTSAXS is its capability of probing very small and negative surface-normal scattering vectors (qz), making it especially well-suited for investigating the structural features of anisotropic materials with a film thickness close to the inverse-surface-normal scattering vector magnitude. This level of precision is not possible with GISAXS alone.

GTSAXS requires a focused beam and an incident angle larger than the critical angle of the substrate in order to minimize refraction effects. It also requires that the illumination occurs near the edge of the sample once the scattering field must exit through it. By carefully controlling these geometrical aspects and collecting suitable scattering images, GTSAXS allows for straightforward analysis using the Born approximation framework, achieving accurate surface scatterer response modelling.

Ultimately, the core novelty of GTSAXS lies in its combination of the advantages of GISAXS and Transmission Small-Angle X-ray Scattering (TSAXS), providing superior capabilities for investigating structural features in nanostructured thin films.cont rg risult's BeauMooMinutesandbased sitsanne UNITEDATURE conference an endeavor explic wchar\#af

    \item \textbf{GPT-3.5 response 2:}
The core novelty of Grazing-Incidence Transmission Small Angle X-ray Scattering (GTSAXS) is its ability to determine the three-dimensional order in nanostructured thin films by probing very small and negative surface-normal scattering vectors. This makes it particularly well-suited for studying structural features of anisotropic materials with a film thickness comparable to the inverse-surface-normal scattering vector magnitude. GTSAXS retains the sensitivity of Grazing-Incidence Small Angle X-ray Scattering (GISAXS) to lateral and perpendicular film structure but eliminates the multiple scattering and refraction issues associated with typical grazing-incidence analysis. By arranging a sample/beam geometry in which the scattering exits through the edge of the substrate and recording scattering images described by straightforward scattering models, GTSAXS offers a unique and robust approach to surface X-ray scattering.

    \item \textbf{GPT-3.5 response 3:}
The core novelty of GTSAXS, or grazing-incidence transmission small-angle X-ray scattering, is its ability to probe very small and negative surface-normal scattering vectors, meaning it can investigate structural features of nanostructured thin films with a comparable film thickness to the inverse-surface-normal scattering vector magnitude. It combines the benefits of fast measurements and film-normal information from grazing-incidence small-angle X-ray scattering (GISAXS), and scattering unperturbed by complicated corrections from transmission small-angle X-ray scattering (TSAXS). GTSAXS eliminates the multiple scattering and refraction issues that are typically associated with grazing-incidence analysis methods, making it well-suited for investigating anisotropic materials and nanostructures. It requires a focused X-ray beam that illuminates the region near the edge of the sample, where the scattering is measured. Importantly, GTSAXS retains sensitivity to the parallel and perpendicular film structures while reducing distortions in the scattering patterns, allowing for analysis within the Born approximation framework and enabling more accurate characterization of nanostructured thin films. Source: [Lu, Ocko, et al. "Grazing-incidence transmission X-ray scattering: surface scattering in the Born approximation"]

\end{itemize}

\newpage
\section{LLM Task Performance}

\begin{figure}[ht]
\centering
  \includegraphics[width=10.0cm]{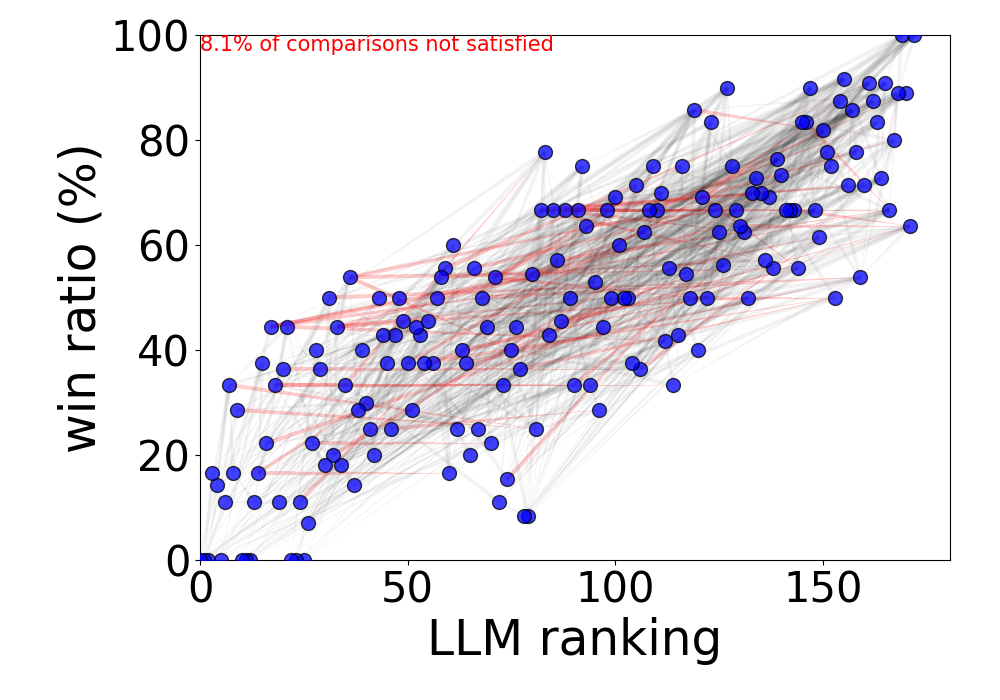}
  \caption{The LLM (OpenAI GPT 3.5) is used to rank documents by ``potential for scientific impact'', using pairwise comparisons where the LLM judges the impact of two scientific documents. The LLM has access to the article text (title, abstract, main text) but no ancillary information such as the name of the journal the paper was published in. The pairwise comparisons are performed on a random set of connections. We ensure that every publication has undergone at least one comparison, but do not compute a dense set of all possible comparisons (818 comparisons, out of a total possible $176^2=30,976$). Using the pairwise comparisons, we then sort the articles into a ranking from lowest impact to highest impact. The sorting is performed by starting with a random order, and then iteratively considering pairs of articles (we iterate both through the current list order, and through the list of comparisons) and accepting a swap if it decreases the total number of misordered pairs. This procedure gradually decreases the fraction of elements that are misordered relative to each other. This fraction does not decrease to zero because there is no guarantee that the pairwise evaluations form a perfectly consistent ordering (viewed as a directed graph, there are cycles in the graph). This sorting yields an ordering where only $8.1\%$ of comparisons are misordered. The graph compares overall win ratio (percentage of time a given document was deemed ``higher impact'' in pairwise comparisons) and uses connecting lines to show the direction of comparison (red lines denote misordered comparisons that could not be satisfied).
}
  \label{fgr:xray_images}
\end{figure}

\begin{figure}[ht]
\centering
  \includegraphics[width=10.0cm]{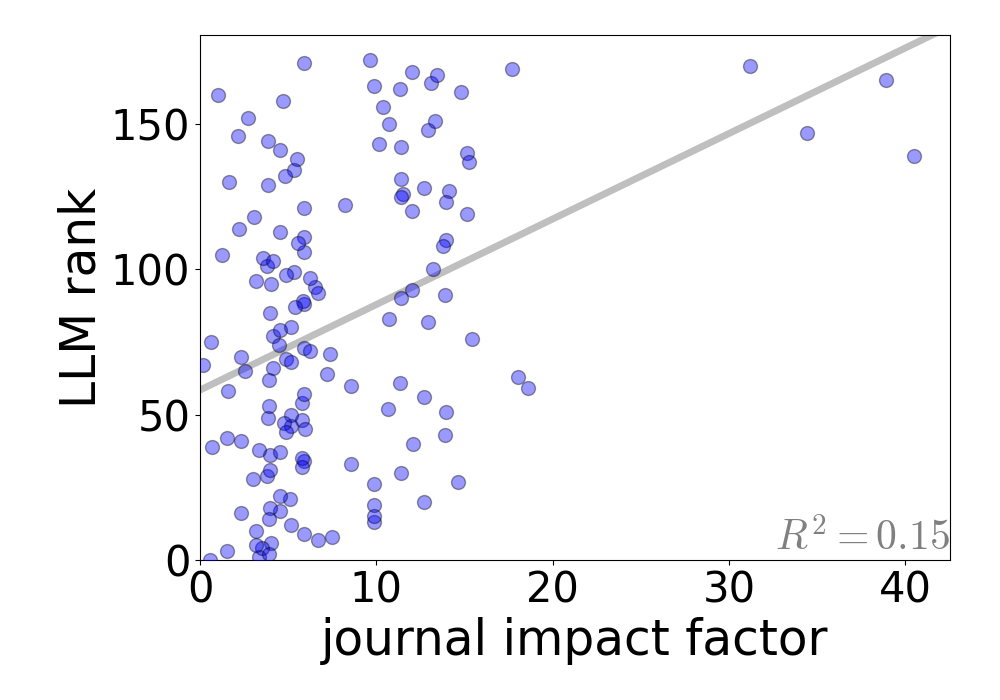}
  \caption{The LLM ranking of publications (by potential for impact) is compared against the impact factor of the journal the manuscript was published in. There is, broadly speaking, agreement between the ordering of publications by LLM assessment and the impact factor. For instance, the highest impact journal articles are indeed rank among the highest by the LLM. Of course, perfect agreement is not expected, since impact assessment is inherently imprecise and subjective; moreover journal impact factor is known to be a coarse proxy for scientific impact. The coefficient of determination for a linear fit to the data ($R^{2}\approx0.15$) suggests some measure of positive correlation between these metrics. Note for the given dataset even perfect sorting by impact factor would not yield perfect correlation (but rather $R^{2}\approx0.69$) since ranking is a contiguous integer list while impact factor is a continuous variable with a non-uniform distribution.
}
  \label{fgr:xray_images}
\end{figure}

\newpage
\clearpage

\begin{table}[h]
\centering
{\small
\setlength{\tabcolsep}{3pt}

\begin{tabular}{c|cccccc|ccc}
\hline
\textbf{Ground truth} & \multicolumn{6}{c|}{ \textbf{LLM assignment} } & \multicolumn{3}{c}{\textbf{metrics}} \\
\hline
 & \small{Self-assembly} & \small{Materials} & \small{Scattering} & \small{Machine-learning} & \small{Photo-responsive} & \small{Other} & \textit{Pr} & \textit{Re} & \textit{Ac} \\
\hline
 Self-assembly & \cellcolor{goodgreen} \textbf{60} & 2 & \textcolor{lightgrey}{0} & \textcolor{lightgrey}{0} & \textcolor{lightgrey}{0} & 1 & 79\% & 95\% & 89\% \\
\hline
 Materials & 16 & \cellcolor{goodgreen} \textbf{31} & 10 & 1 & \textcolor{lightgrey}{0} & \textcolor{lightgrey}{0} & 84\% & 53\% & 81\% \\
\hline
 Scattering & \textcolor{lightgrey}{0} & \textcolor{lightgrey}{0} & \cellcolor{goodgreen} \textbf{11} & \textcolor{lightgrey}{0} & \textcolor{lightgrey}{0} & \textcolor{lightgrey}{0} & 41\% & 100\% & 91\% \\
\hline
 Machine-learning & \textcolor{lightgrey}{0} & 1 & 5 & \cellcolor{goodgreen} \textbf{16} & \textcolor{lightgrey}{0} & 1 & 94\% & 70\% & 95\% \\
\hline
 Photo-responsive & \textcolor{lightgrey}{0} & 1 & \textcolor{lightgrey}{0} & \textcolor{lightgrey}{0} & \cellcolor{goodgreen} \textbf{10} & \textcolor{lightgrey}{0} & 100\% & 91\% & 99\% \\
\hline
 Other & \textcolor{lightgrey}{0} & 2 & 1 & \textcolor{lightgrey}{0} & \textcolor{lightgrey}{0} & \cellcolor{goodgreen} \textbf{2} & 50\% & 40\% & 97\% \\
\hline
\end{tabular}
}
\caption{The corpus of scientific documents were manually classified by a human into 6 thematic categories. The table shows the number of publications sorted into these categories by LLM (OpenAI GPT 3.5). From these one can compute the true positive (TP), true negative (TN), false positive (FP), and false negative (FN) counts. We compute additional metrics: precision, $Pr = \mathrm{TP}/(\mathrm{TP}+\mathrm{FP})$; recall, $Re=\mathrm{TP}/(\mathrm{TP}+\mathrm{FN})$; and accuracy, $Ac=(\mathrm{TP}+\mathrm{TN})/(\mathrm{TP}+\mathrm{FP}+\mathrm{FN}+\mathrm{TN})$. Overall, the LLM is highly successful at this imprecise task (accuracy 81-99\%), with the majority of errors being reasonable (e.g. ambiguous classification between \textit{materials} category, or \textit{self-assembly} category more specifically).}
\end{table}

\newpage
\section{Examples of Image Retrieval}

\begin{figure}[ht]
\centering
  \includegraphics[width=15.0cm]{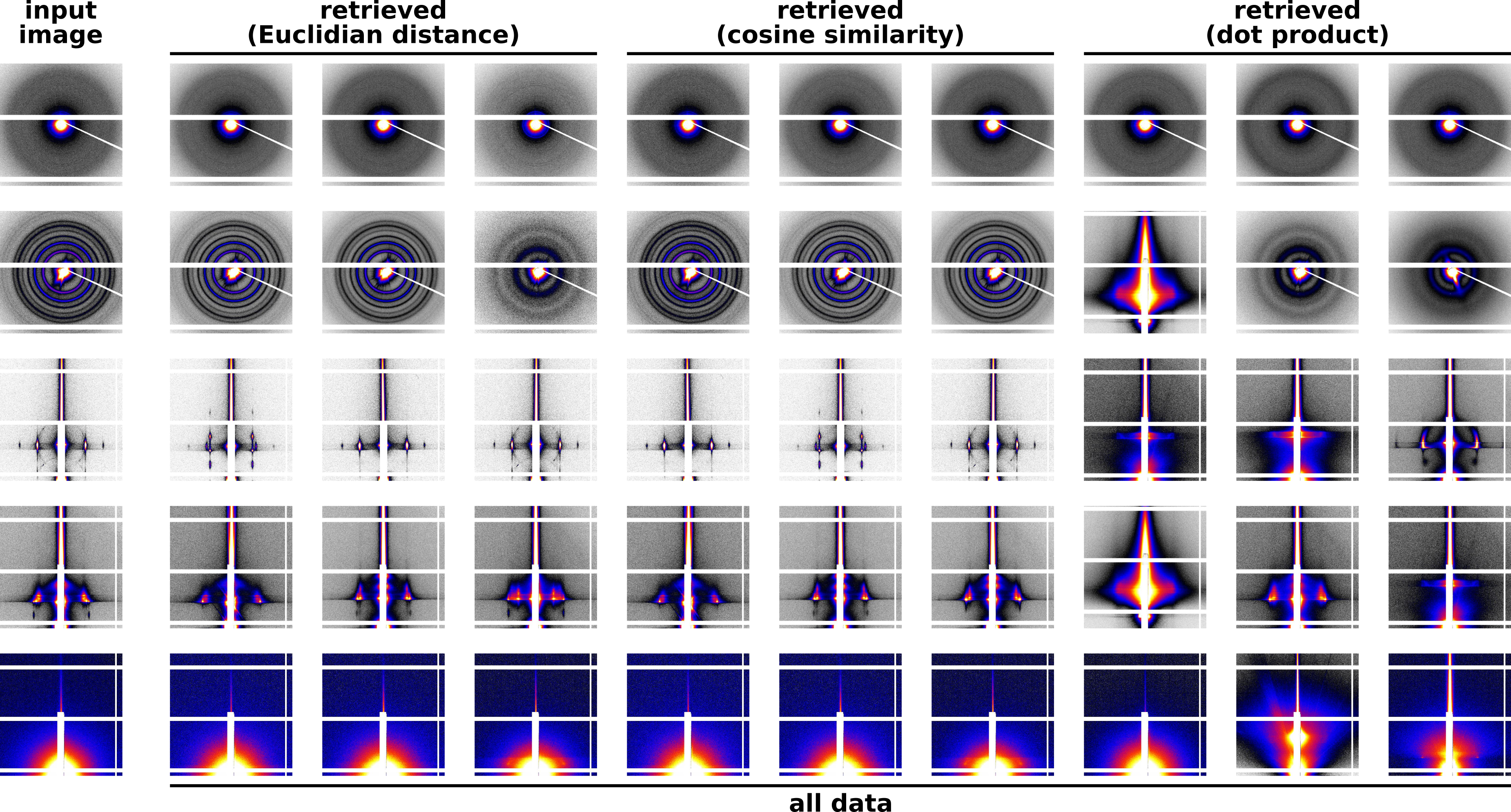}
  \caption{Examples of image retrieval from a database of $50,923$ images. Input images are small-angle x-ray scattering (SAXS) detector images, including grazing-incidence (GISAXS) data, collected at the Complex Materials Scattering (CMS, 11-BM) beamline at the National Synchrotron Light Source II (NSLS-II). Examples are provided for Euclidian distance (which measures how close in meaning the images are), cosine similarity (which measures how similar in theme or topic the images are), and dot product (which measures overlap in the underlying concepts). Retrieved images show meaningful similarity.
}
  \label{fgr:xray_images}
\end{figure}

\begin{figure}[ht]
\centering
  \includegraphics[width=15.0cm]{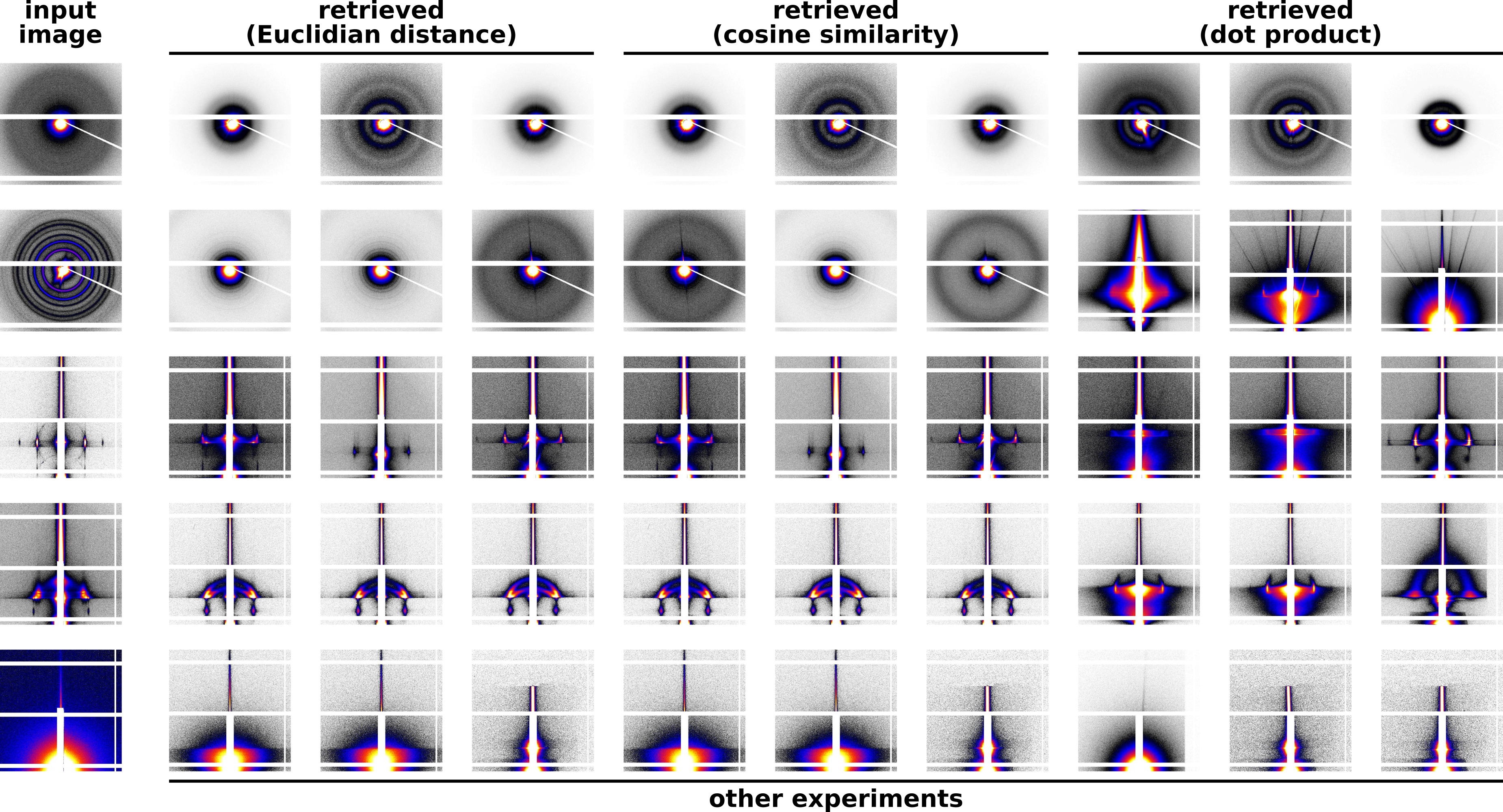}
  \caption{Examples of image retrieval for SAXS/GISAXS inputs, where images from the same beamline experiment as the input were excluded. This demonstrates the ability to discover similar (conceptually related) data in different experiments (or even different materials).
}
  \label{fgr:xray_images2}
\end{figure}